\documentclass[preprint,5p,times,twocolumn]{elsarticle}

\usepackage[ruled]{algorithm2e}
\SetAlFnt{\small}
\SetAlCapFnt{\small}
\SetAlCapNameFnt{\small}
\SetAlCapHSkip{0pt}
\IncMargin{-\parindent}

\usepackage{booktabs}
\usepackage{multirow}
\usepackage{multicol}
\usepackage{listings}
\usepackage{amssymb}
\usepackage{amsfonts}
\usepackage{amsmath}
\usepackage{graphicx,rotating}
\usepackage{url}
\usepackage{mathtools,caption,theorem}
\usepackage{dblfloatfix}
\usepackage{balance}
\usepackage[english]{babel}

\journal{Applied Soft Computing Journal}

\begin{document}

\begin{frontmatter}



\title{Optimizing groups of colluding strong attackers in mobile urban communication networks with evolutionary algorithms\tnoteref{fastrack}}

\tnotetext[fastrack]{This paper is an extended, improved version of the paper \emph{Black Holes and Revelations: Using Evolutionary Algorithms to Uncover Vulnerabilities in Disruption-Tolerant Networks} presented at EvoComNet2015 and published in: Applications of Evolutionary Computing, Proceedings of 18th European Conference, EvoApplications 2015, Copenhagen, Denmark, April 8-10, 2015, LNCS 9028, pp. 29-41, Springer, 2015.}

\author[l1]{Doina Bucur}
\ead{d.bucur@rug.nl}
\author[l2]{Giovanni Iacca}
\ead{giovanniiacca@incas3.eu}
\author[l3]{Marco Gaudesi}
\ead{marco.gaudesi@polito.it}
\author[l3]{Giovanni Squillero}
\ead{giovanni.squillero@polito.it}
\author[l4]{Alberto Tonda}
\ead{alberto.tonda@grignon.inra.fr}
\address[l1]{Johann Bernoulli Institute, University of Groningen, Nijenborgh 9, 9747 AG Groningen, The Netherlands}
\address[l2]{INCAS\textsuperscript{3}, Dr. Nassaulaan 9, 9401 HJ, Assen, The Netherlands}
\address[l3]{Politecnico di Torino, Corso Duca degli Abruzzi 24, 10129, Torino, Italy}
\address[l4]{INRA UMR 782 GMPA, 1 Avenue Lucien Br\'{e}tigni\`{e}res, 78850, Thiverval-Grignon, France}

\begin{abstract}
In novel forms of the Social Internet of Things, any mobile user within communication range may help routing messages for another user in the network. The resulting message delivery rate depends both on the users' mobility patterns and the message load in the network. This new type of configuration, however, poses new challenges to security, amongst them, assessing the effect that a group of colluding malicious participants can have on the global message delivery rate in such a network is far from trivial. In this work, after modeling such a question as an optimization problem, we are able to find quite interesting results by coupling a network simulator with an evolutionary algorithm. The chosen algorithm is specifically designed to solve problems whose solutions can be decomposed into parts sharing the same structure. We demonstrate the effectiveness of the proposed approach on two medium-sized Delay-Tolerant Networks, realistically simulated in the urban contexts of two cities with very different route topology: Venice and San Francisco. In all experiments, our methodology produces attack patterns that greatly lower network performance with respect to previous studies on the subject, as the evolutionary core is able to exploit the specific weaknesses of each target configuration.
\end{abstract}


\end{frontmatter}


\section{Introduction}
\label{sec:intro}

The so-called \emph{Social Internet of Things} calls for nearly ubiquitous communicating devices. There is today a need to integrate low-cost, low-power devices to support networking services in more effective and efficient ways. 
In such a scenario, new solutions are continuously developed and deployed, while approaches that just a few decades ago were used only in highly complex, niche applications
are now literally brought down to earth --- Delay-Tolerant Networks (DTNs) are a technology originally developed for space communications that, over the years, made its way down to quite mundane applications~\cite{fall2008dtn}. 

Emerging technologies and applications are posing serious problems to designers. In most cases there is not enough time to thoroughly validate them, or even to simply analyze their possible failures and problems. Engineers are forced to resort to their experience to choose heuristics that look reasonable, and then observe the actual outcome from real applications. Security in DTNs is a paradigmatic case: such networks need to remain open to all willing participants, and few malicious participants may try to disrupt communications, for instance, routing no messages to other nodes or injecting large number of messages into the network. While such a risk is plausible, precisely assessing DTNs' vulnerabilities is hard.

This paper focuses precisely on evaluating the amount of damage that can be caused to a DTN by a group of synchronized attackers with deep knowledge about the network. Given a scenario, we propose to optimize attackers for minimizing the performances of the network using a heuristic methodology.
It is important to note that the adoption of such methodology is more a necessity than a choice: determining the most effective attack for a given network was proven to be NP-hard~\cite{MaxProp}, the complexity and number of variables involved in the problem preclude the use of formal techniques and the size of the scenarios prevent exhaustive analyses.

The idea of using heuristic methods to disprove a property of a system when formally proving it is not possible, is not a novelty in itself. The simplest approach, namely random sampling of the parameter space, is often used under the assumption that the effort employed failing to find a counter example may be sensibly linked to the degree of confidence that a counter example does not actually exist.

Repeated random sampling has also be used as a means to estimate numerical quantities when complexity and dimensionality of a problem impedes the application of analytic analyses. In the specific case of DTNs performance, it has been considered in~\cite{Burgess:Surviving}, although limited only to small networks and attackers with no information about the environment. However, random sampling is unlikely to provide any interesting result when the goal is to detect a very specific corner-case scenario, such as the damage caused by specialized attackers that are fully aware of the network characteristics. Finally, when the search space is too vast, even the effort required to get a significant sampling could be excessive.

In this work, we move forward from random sampling by using an evolutionary algorithm (EA) to \emph{optimize the attackers' parameters} in order to inflict the maximum possible damage to the network. We overcome the limitations of random sampling by using the capability of the EA to drive random search towards specific regions of large search spaces. Furthermore, we extend the features of a classical evolutionary algorithm to enable it to find a \emph{team of colluding attackers}. As the members of such a team cooperate in order to maximize the cumulative damage, even at the expense of the damage caused by each single attacker, the approach is a form of \emph{cooperative co-evolution}, an open area of research for which very few successful strategies have been found so far.

We tested the proposed methodology on medium-sized networks describing urban scenarios with different topologies, where a number of agents, i.e., the network nodes, move realistically. The results clearly demonstrate the efficacy of the approach: we found scenarios where even few (up to $10\%$ of the total network size), highly optimized attackers can reduce the global data delivery in the network by over 90\%, when compared to the network with no attackers. We also observed that the composition of the attacker team obtained by evolution changed when cooperative co-evolution is used, demonstrating that such scheme leads to synergistic solutions not found by classical evolutionary algorithm. 

The rest of the paper is organized as follows: the next section summarizes the research background; Section \ref{sec:method} details the proposed methodology; Section \ref{sec:exp} reports the experimental evaluation; Section \ref{sec:rw} surveys the related work; finally, Section \ref{sec:concl} concludes the paper.

\section{Background}
\label{sec:backgr}

This section first gives an overview of the application domain of Delay-Tolerant Networks. Sections~\ref{sec:background-DTN}-\ref{sec:background-DTN-protocols} describe the paradigm of routing in DTNs, and First Contact, the DTN protocol under study. Section~\ref{sec:background-movement-models} summarizes the mobility model that an urban DTN node follows, from the literature. Section~\ref{sec:background-security} describes the two main types of security attacks relevant: black hole and flooding attacks. Finally, Section~\ref{ssec:2backgr:evolutionary_computation} gives an overview of the EA field.

\subsection{Delay-Tolerant Networks: performance objectives}
\label{sec:background-DTN}

Delay-Tolerant Networking was designed to cater for \emph{message routing} in practical applications with \emph{heavy node mobility}. In such applications, the connectivity pattern between nodes in the network can be either predictable or unpredictable with time. An example DTN with predictable connectivity is that of a mixed terrestrial-and-space network where some of the nodes are Low-Earth Orbiting Satellites, and the rest are ground users; this was the application for which DTN-specific routing protocols were originally designed~\cite{Jain:SIGCOMM04}. More recently, DTNs have also been proposed in scenarios with nearly unpredictable connectivity. This is the case of animal-tracking applications~\cite{DTN:animal} and opportunistic urban networks. An example of the latter is the 30-bus experimental DieselNet~\cite{MaxProp}, in which urban vehicles constrained to city roads act as mobile message routers. It is urban scenarios with unpredictable connectivity that we study in this paper.

Given an application scenario, the main \emph{performance factors} for a DTN message-routing protocol quantify the protocol's ability to route messages in that scenario, and the timeliness of the routing:

\begin{description}
	\item[Delivery rate] The percentage of messages injected in the network by nodes which were successfully delivered to their destination nodes.
	\item[Message delay] The average time interval between a message injection in the network until its delivery.
\end{description}

\subsection{DTN routing: the First Contact protocol}
\label{sec:background-DTN-protocols}

A DTN routing protocol essentially implements a logic to achieve message routing in the mobile network, end-to-end from the source of a message to its destination, over a connectivity graph which varies in time and is by nature disconnected. Given these scenarios, the protocol logic cannot be based on standard distributed algorithms for computing shortest paths end-to-end in a graph: the routing can not converge on correct routes when the network graph is highly dynamic. Instead, DTN message communication on a path between source and destination include long-term \emph{storage} of the message in the nodes' (finite) node buffers, until an opportunity for further delivery of the message arises. This ability to safely \emph{delay} the forwarding of a message is typical of DTN routing protocols.

DTN protocol design follows a simple taxonomy based on the following features:

\begin{description}
	\item[Network knowledge] A protocol aiming to compute optimal paths at a node would be helped if the node is able to predict the future network conditions: the pattern of contact with other nodes, the set of nodes with congested buffers, and the pattern of traffic demands. While network knowledge may be acquired in practice by an attacker via monitoring the network, many protocols cannot assume any (i.e., are \emph{zero-knowledge}). 
	\item[Message replication] \emph{Forwarding protocols} simply route the original message through the network. \emph{Replicative protocols} introduce into the network a number of copies of each original message, each of which is then forwarded independently with the aim that at least one copy reaches the destination.
\end{description}

We study here one of the simplest and most common DTN routing protocols, namely \emph{First Contact} (FC)~\cite{Jain:SIGCOMM04}. FC is zero-knowledge and forwarding; it routes messages opportunistically using any available contacts with other nodes. A single copy of each message in the network exists at a time, and it is forwarded to the first available contact (if more contacts are available, one is chosen randomly among all the current contacts). On simple network topologies, FC was shown to have performance comparable to partial-knowledge protocols; this degrades in complex topologies to varying degrees, depending on the network load.

\subsection{Node movement models in DTNs} 
\label{sec:background-movement-models}

Models describing realistically~\cite{ONE} the free, stochastic movement in urban environments of nodes of different kind (pedestrian, cars, buses, etc.) can be used to study an urban DTN computationally. In a DTN simulation, each movement model is associated to a different map layer. A map layer describes the areas of the map reachable by the associated kind of nodes. The building blocks of these node movement patterns are a node's points of interest (POIs) located on a map layer.

Nodes in our DTNs follow a classic movement model: \emph{random waypoint with shortest paths}. In this model, node randomly chooses a destination point from a set of points of interest; travels there at a realistic speed on the shortest path; takes a break. Then, it repeats the process. 
This can be considered realistic: in \cite{Burgess:Surviving}, which evaluated the theoretical resilience of the real-world DieselNet and Haggle DTN prototypes, the network logs showed that Haggle nodes mimicked the random-waypoint model fairly closely.

\subsection{Types of security attacks}
\label{sec:background-security}

Like a DTN routing protocol, an attacker also may or may not have knowledge of the future connectivity patterns in the network. A \emph{strong attacker} has full network knowledge, i.e., will know:
\begin{itemize}
	\item The structure of the city map;
	\item The pattern of network encounters: for example, a statistical estimation of how many honest nodes will be in proximity at any given map location;
	\item The statistical pattern of new messages that honest nodes will forward when encountered, and the statistical pattern of buffer availability at honest nodes.
\end{itemize}

Furthermore, a group of \emph{colluding attackers} has the means necessary to synchronize their individual attacks, rather than executing an independent logic. Any of the colluding nodes may adopt one of the following attack logics:
\begin{description}
	\item[Black hole attacks] The attacker drops a percentage of the packets received: this percentage is 100\% in black hole attacks. 
	\item[Flooding attacks] The attacker executes the same routing protocol as honest nodes, but attempts a denial-of-service procedure by injecting a (large) number of (large) messages into the network. 
\end{description}

\subsection{Evolutionary Computation}
\label{ssec:2backgr:evolutionary_computation}

Evolution is the biological theory that animals and plants have their origin in other types, and that the distinguishable differences are due to modifications in successive generations. Natural evolution is based on random variations, but it is not a random process: variations are rejected or preserved according to objective evaluations, and only changes that are beneficial to the individuals are likely to spread into subsequent generations. Darwin called this principle ``natural selection'' \cite{Darwin1859}: a deterministic process where random variations ``afford materials''.

When natural selection causes variations to be accumulated in one specific direction the result may strikingly resemble a deliberate optimization process. However, such optimization processes only required to assess the effect of random changes and not the ability to design intelligent modifications. Several scholars were inspired by such an outcome and tried to reproduce the process for solving practical optimization problems in various application domains, while others tried to mimic it to better understand its underlying mechanisms.

\emph{Evolutionary Computation} (EC) is the offshoot of computer science focusing on algorithms loosely inspired by the theory of evolution. The definition is deliberately vague since the boundaries of the field are not, and cannot be, sharply defined. EC is a branch of \emph{computational intelligence}, and it is also included into the broad framework of \emph{bio-inspired meta-heuristics}. EC does not have a single recognizable origin. Some scholars identify its starting point in 1950, when Alan Turing drew attention to the similarities between learning and evolution \cite{Turing1950}. Others pointed out the inspiring ideas that appeared later in the decade, despite the fact that the lack of computational power impaired their diffusion in the broader scientific community \cite{Fogel1998}. More commonly, the birth of EC is set in the 1960s with the appearance of three independent research lines: John Holland's \emph{genetic algorithms} \cite{Holland1975}; Lawrence Fogel's \emph{evolutionary programming} \cite{Fogel1962}; Ingo Rechenberg's and Hans-Paul Schwefel's \emph{evolution strategies} \cite{Beyer2002}. The three paradigms monopolized the field until the 1990s, when John Koza entered the arena with \emph{genetic programming} \cite{Koza1992}. Nowadays, all these methods, together with several variants proposed over the years, have been grouped under the umbrella term of \emph{evolutionary algorithms} (EAs). 

\begin{figure}[htp]
\centering
\includegraphics[trim=0cm 17.5cm 12cm 0cm, clip=true, width=0.95\columnwidth]{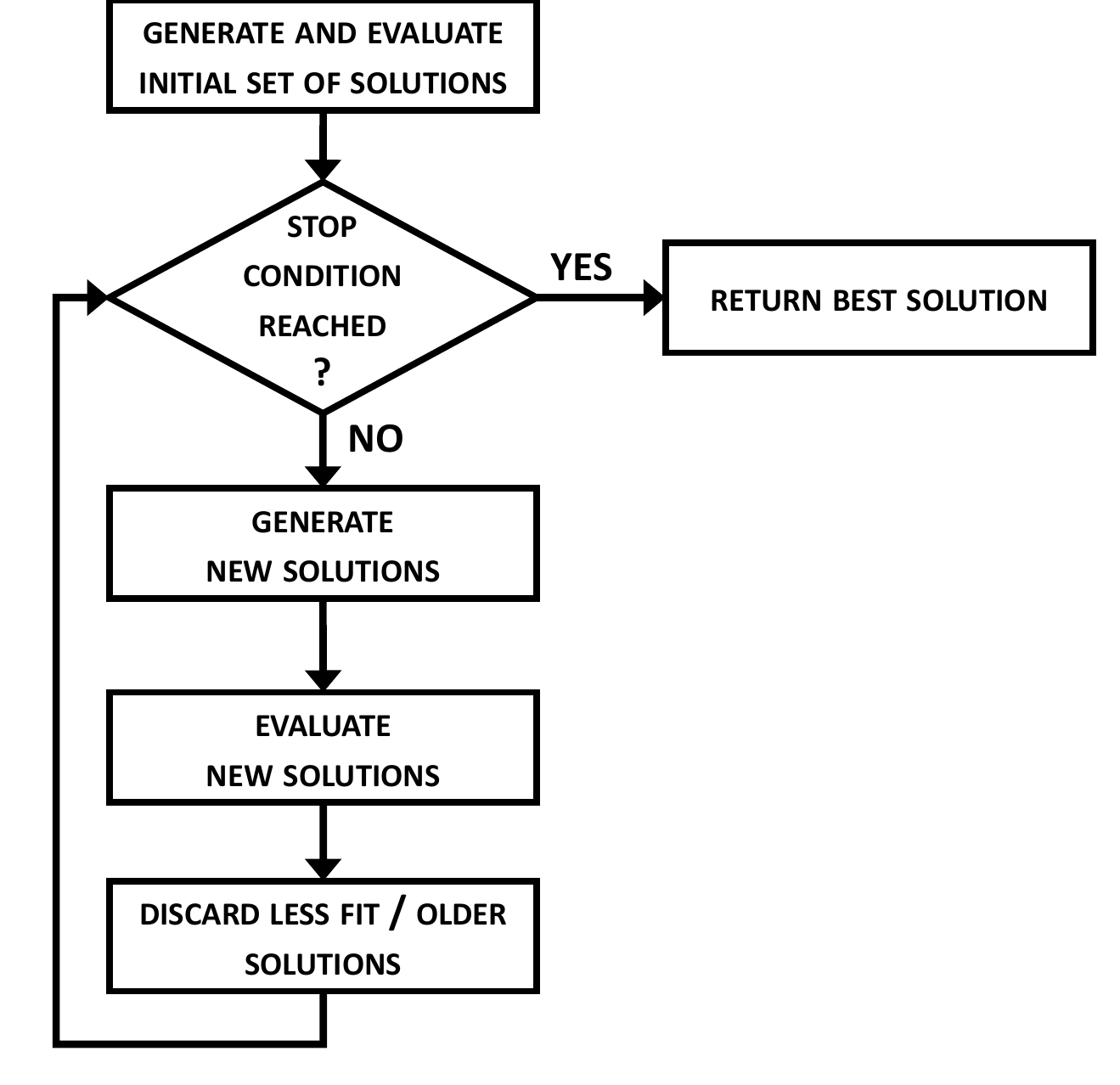}
\caption{Flowchart of an evolutionary algorithm}
\label{fig:ea-general-flowchart}
\end{figure}

When EAs are used to solve a specific problem, i.e. optimize solutions for it, an \emph{individual} is a single candidate solution, and its \emph{fitness} is a measure of its capacity of solving the problem; the set of all candidate solutions that exists at a particular time represents the \emph{population}. Evolution proceeds through discrete steps called \emph{generations}. In each of them, the population is first expanded and then collapsed, mimicking the processes of breeding and struggling for survival (Figure~\ref{fig:ea-general-flowchart}).

Usually, parents are chosen for breeding stochastically, with the best candidate solutions having higher probabilities to generate offspring. As a result, new candidate solutions are more likely to inherit favorable traits. Conversely, the removal of individuals is usually deterministic: the less fit, and possibly the oldest ones, are deleted.
The mechanisms used to generate the offspring are collectively named \emph{genetic operators}. They can be divided into \emph{recombinations} and \emph{mutations}: the former methods mix together the information contained in two or more solutions to create new ones; the latter ones work by changing the structure of a single solution. Recombination operators are able to coalesce good characteristics from different solutions, and provide a very effective mechanism to explore the search space; mutation operators, on the other hand, allow the fine-tuning of the candidate solutions. Maintaining a set of solutions, EAs are resilient to the attraction of local optima \cite{Eiben2010}. 

Over the years, EAs were proven capable to solve quite difficult problems with very complex fitness landscapes, including open problems related to networking and protocols. In particular, EAs are known to greatly outperform random sampling for case studies where classical optimization techniques are not viable \cite{Eiben2010}. Evolutionary optimizers have been successfully exploited both in stationary and dynamic situations, and they were demonstrated able to identify either single optima or Pareto sets in multi-objective problems, see Section~\ref{sec:rw} for a summary of relevant prior work.


\section{Related Work}
\label{sec:rw}

\subsection{Random Sampling}
\label{ssec:4rw:randomly_testing_weak_attacks}

To the best of our knowledge, the only experimental work on the assessment of DTN robustness was performed by Burgess et al.\cite{Burgess:Surviving}, who evaluated the resilience of DTNs of 30 nodes when running four protocols. Weak attacks (without free range, but with attackers forced to use the routes previously used by the honest nodes) were simulated by randomly reassigning some of the honest nodes as attackers. 

\subsection{Greedy Heuristic}
\label{ssec:4rw:attack_heuristics_to_compute_strong_attacks}

To generate strong attacks (again, without free range), the same authors~\cite{Burgess:Surviving} used a greedy heuristic in response to the intractability of the vertex vulnerability problem. First, instead of aiming to minimize the delivery rate of the network, the simpler quality-of-service metric of \emph{total reachability} substitutes it, which in turn can be minimized effectively with heuristics. This metric is defined as follows. A DTN $D = (N,C)$ is a single \emph{predefined list of connection events} $C$ on the set of honest nodes $N$ of size $n$; this can be obtained by monitoring the long-term execution of a real-world DTN. Two nodes are temporally connected in $D$ if there exists a (temporally non-decreasing) sequence of connection events in $C$. Then, the total reachability of $D$, denoted $R(D)$, is the number of pairs of temporally connected nodes (excluding reflexive pairs). To select $k$ attackers out of the set $N$ while minimizing $R(D)$ on a predefined $D$, the \emph{greedy heuristic} simply selects as the \emph{i}th attacker that node which lowers the total reachability of $D$ (excluding the first $i-1$ attacker) the most. While this greedy heuristic is not proven analytically to be optimal at minimizing $R(D)$, it is shown experimentally to give similar results as a brute-force method, for $k\le 5$, on the two DTNs under study. 

The greedy heuristic is then shown to outperform a random selection of nodes when $k \ge \frac{n}{10}$ (i.e., at least 3 strong attackers for $n=30$) on the DieselNet DTN, and for any $k\ge 1$ on the Haggle DTN, for black hole attacks and the MaxProp routing protocol. The heuristic is particularly advantageous for a very large $k$; there, the difference between the delivery rate found when $k = \frac{n}{2}$ is $30-40\%$ lower on the two DTNs than with random node selection.

\subsection{Bio-inspired Heuristics}

Evolutionary computation has been demonstrated a powerful means for tackling the inherent complexity of networking. In \cite{Baldi:GA}, Baldi et al. proposed the use of a genetic algorithm for minimizing the performance of a network, with the final goal to pinpoint potential bottlenecks of the topology. The approach exploits an accurate network simulator coupled with a rudimentary evolutionary optimizer written in Perl. It could be considered a proof of concept, as experimental results do not show any real application.

A feasibility study of the approach described in the present work, exploiting a far more conventional EA and tested only on a reduced set of scenarios, was presented in \cite{Bucur:EvoApp15}. In previous research, we showed the efficacy of an evolutionary algorithm also in the context of wireless sensor networks (WSN), for which we tested two kinds of collection tree protocols \cite{Bucur:ASOC14,Bucur:AdHoc15}. We tackled different network topologies composed of up to 50 nodes, making use of a real-code simulator, which enables to analyze the complete software implementation of the protocols under analysis. Furthermore, the gathered data enabled us to pinpoint a set of topological factors which correlate with extreme traffic under collection routing. In \cite{Bucur:GECCO14}, we further enhanced such analysis through the use of a multi-objective evolutionary algorithm, in the attempt to explore the WSN search space from a multi-objective perspective rather than using lexicographic order of fitness functions.

\section{Proposed Methodology}
\label{sec:method}

As stated in the introduction, the core idea of this work is to expose vulnerabilities in a DTN by devising a set of effective attackers. As a single attacker is unlikely to be able to damage a DTN network, we seek for teams of colluding malicious nodes. Both the number of attackers and their type (black hole or flooding) should be optimized in order to create the maximum damage. Attackers are optimized using an advanced evolutionary algorithm that is based on an recent cooperative co-evolution approach that is not only able to optimize the parameters of each individual attacker, but also to optimize the composition of the team of colluding attackers.


In more detail, we simulate a realistic DTN over an urban environment defined by a topological map and a set of POIs. We assume two type of nodes: \emph{honest} and \emph{malicious}. As discussed earlier, for each honest node $i$ the predetermined moving path $P^{H}_{i}$ is a sequence of random points of interest. For added realism, a small number of these points, such as main tourist attractions, may be given a higher probability of being selected as next destination. On the contrary, for each malicious node $i$, the path $P^{M}_{i}$ is a sequence of points of interest chosen by the evolutionary optimizer to cause maximum damage in the network. Honest nodes execute the FC routing protocol, while malicious nodes can act either as data flooders or black holes.

In the proposed framework, we use the Opportunistic Network Environment simulator (The ONE) \cite{ONE}\footnote{The tool is available at \url{http://www.netlab.tkk.fi/tutkimus/dtn/theone/}.} coupled with the evolutionary toolkit $\mu$GP \cite{Sanchez2011}\footnote{The tool is available at \url{http://ugp3.sourceforge.net}.}. The reasons for using $\mu$GP are manifold: first, the design of this framework is based on the notion of an \emph{external evaluator}, which simplifies the integration with an external network simulator; secondly, the algorithm available in $\mu$GP features a built-in support for multiple fitness functions, that can be evaluated both in a lexicographical order and in a multi-objective approach; then, the evolutionary engine available in $\mu$GP makes use of self-adaptation techniques, greatly limiting the number of parameters that require to be set. Finally, $\mu$GP provides both a classical EA and a cooperative co-evolution scheme called \emph{Group Evolution} (see below).

\begin{figure}[htb]%
\centering
\includegraphics[trim=2cm 4cm 1cm 2cm, clip=true, width=0.95\columnwidth]{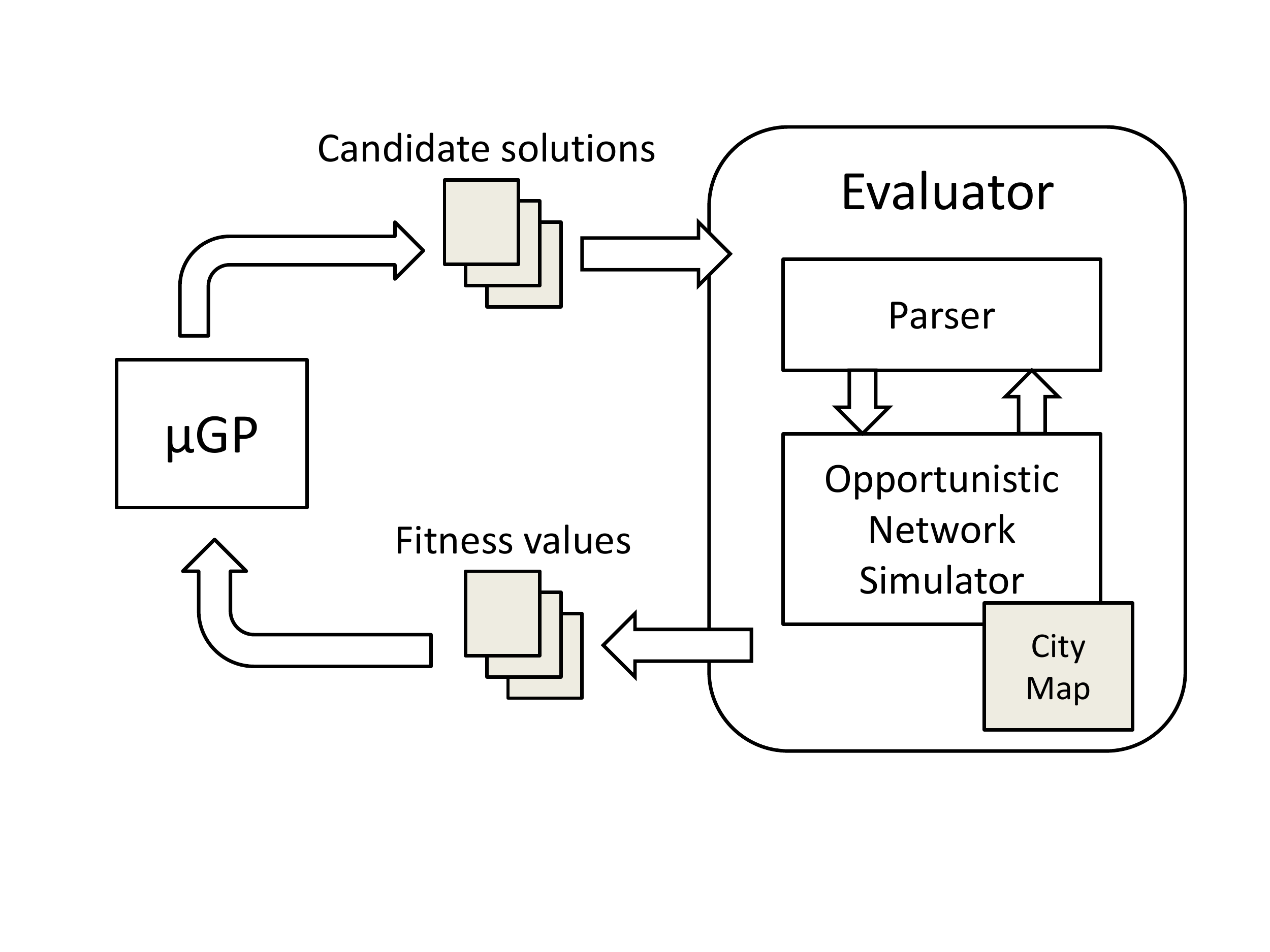}%
\caption{Structure of the proposed framework. Candidate solutions and fitness values are internally represented as text files.}%
\label{fig:framework}%
\end{figure}

The resulting evolutionary optimization process, depicted in Figure \ref{fig:framework}, can then be summarized as follows: given a DTN of $N$ total nodes, and any parameters of the urban environment, find a group of attackers of size $k<N$, each one with its peculiar movement patterns $P^{M}_{i} (i=1\ldots k$) and its characteristics (movement model and attack type) which would lower the data delivery rate (DDR) of the DTN the most, while maximizing also its average latency. 

The following section gives details about the evolutionary core and the Group Evolution scheme, while the internal solution representation and the fitness function definitions are described in Section \ref{ssec:5method:encoding} and \ref{ssec:5method:fitness}, respectively.

\subsection{Evolutionary core}
\label{ssec:5method:evolutionary_core}

A noticeable branch of EC is cooperative co-evolution (CCE), that is, broadly speaking, the study of evolutionary algorithms whose final goal is achieved by a solution composed of different sub-solutions that cooperates to reach the common goal. The idea of CCE dates back to the origin of EC, yet its inherent problems are far from being solved: important contributions are appearing regularly in the scientific literature (e.g., \cite{Potter1994,Dunn2006,Thomason2008,Waibel2009,tonda2011lamps}). In the last decade, the CCE popularity further boasted due to robotics applications where teams of robots can be asked to perform collective tasks \cite{Panait2005}. 

In CCE, sub-solutions may be heterogeneous or homogeneous, and combining them might be more or less trivial. Nevertheless, almost all approaches strive to optimize the single parts independently, while trying periodically to group them into an effective set, possibly exploiting heuristics or ad-hoc tweaks. One of the main challenges in CCE is that optimizing a single component may not be beneficial to the global solution, yet the algorithm has to harmonize the two possibly contrasting selective pressures.

\begin{figure}[hbt]%
\includegraphics[trim=2cm 4cm 2cm 3cm, clip=true, width=\columnwidth]{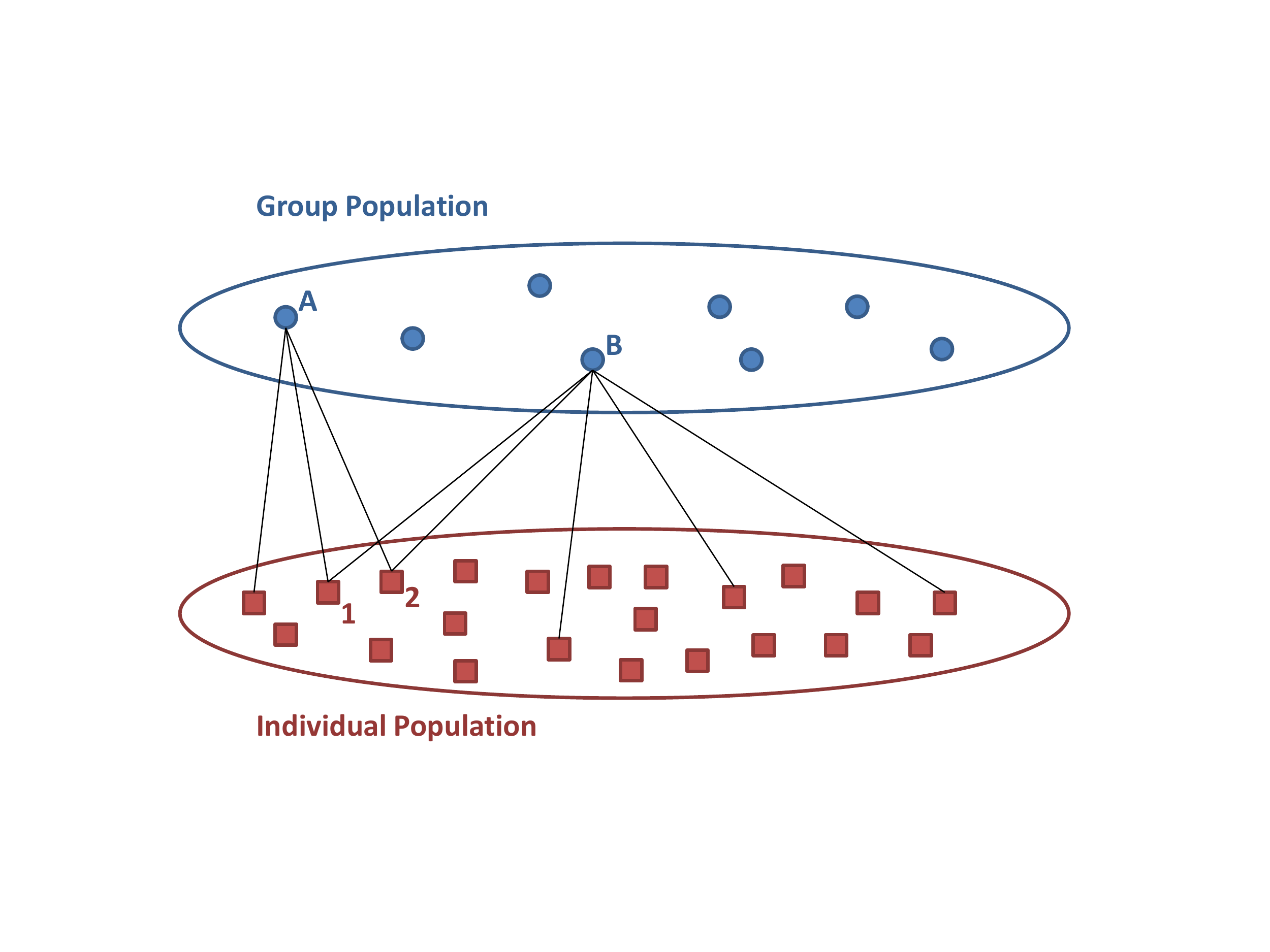}%
\caption{A high-level scheme of the two-population approach used by GE. Groups are sets of individuals taken from the individual population. The same individual can appear multiple times in the same group or in different groups: for example, individuals $1$ and $2$ belong to both groups $A$ and $B$.}%
\label{fig:group-evolution}%
\end{figure}

\emph{Group evolution} (GE) is yet another take on CCE, natively provided by $\mu$GP. In GE, the individual optimization phase and the group optimization phase are blended into a single seamless process \cite{Sanchez2011a}. Individuals are merely the parts that can be assembled to compose the groups, while groups are the actual candidate solutions. GE stores a population of individuals and a separate population of groups (see Figure \ref{fig:group-evolution}), but new individuals and new groups are created with no predefined order: the evolutionary core may choose the best sequence of operators acting on individuals, and operators acting on groups. However the user may still impose a minimum or maximum cardinality for groups. 

Another peculiarity of GE is that single individuals and sets of individuals (i.e., groups) are evaluated by the very same objective function (e.g., the loss of performance in a DTN in the context of this paper). This choice enables both a generalization in the fitness calculation and a tighter integration between the two levels of evolution. The fitness assigned to groups depends on the cumulative effect of the individuals belonging to it, while the contribution of each single individual is also stored and used during comparison.

More operatively, each group is a set of references to individuals in the individual population: so, the same individual can belong simultaneously to multiple groups. At every generation, once new groups and individuals are created and evaluated, groups are sorted by their fitness function, and the worst are removed, factually deleting references to certain individuals. After this process, \emph{orphans}, i.e., individuals not belonging to any group in the current group population, are also deleted.

\begin{figure}[!ht]
\centering
\includegraphics[trim=0cm 9cm 12cm 0cm, clip=true, width=0.95\columnwidth]{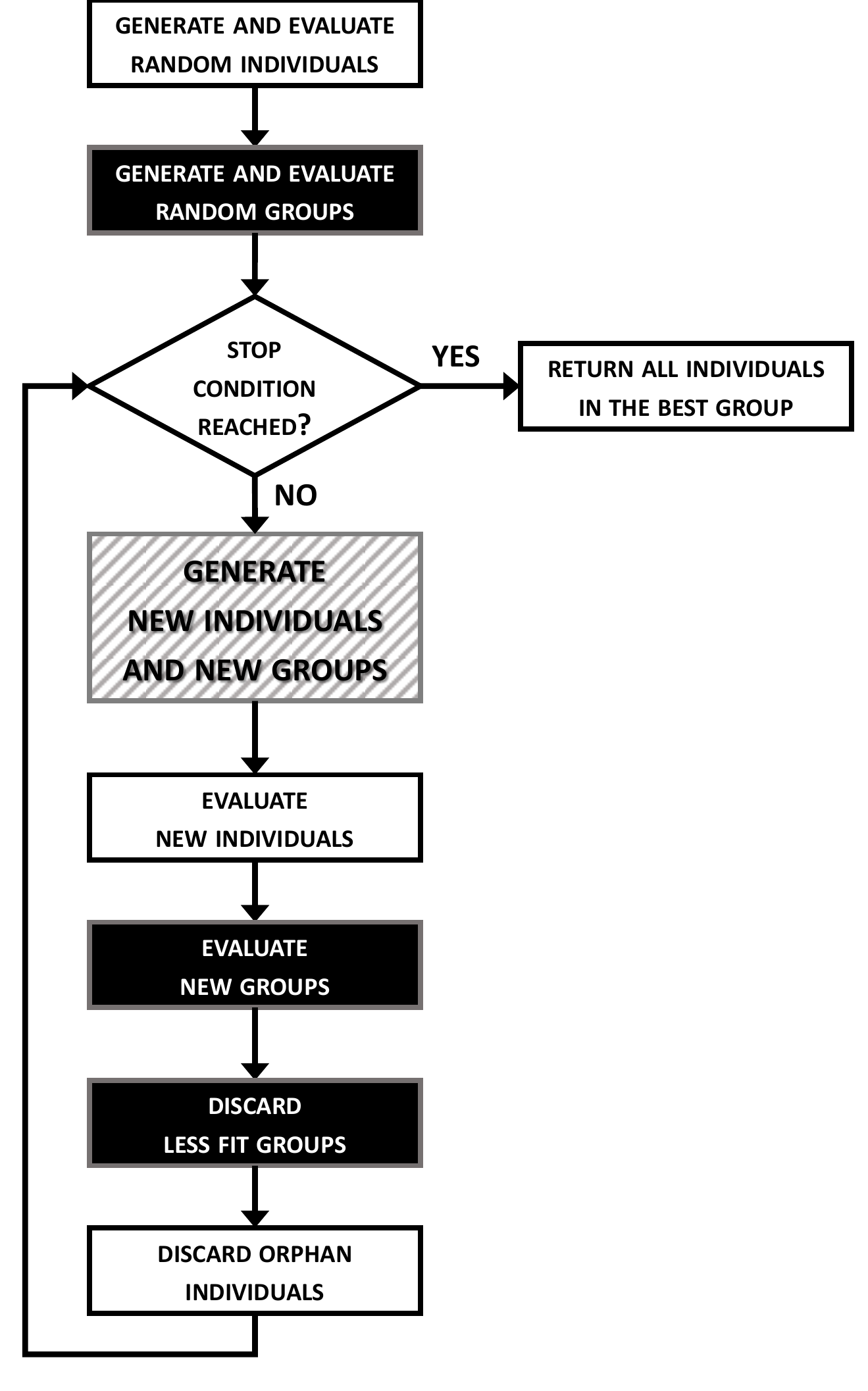}
\caption{Flowchart of an evolutionary algorithm using group evolution. Operations exclusive to GE are depicted in black. New groups and new individuals are created in a single uniform step, while to evaluate new groups the evaluation of new individuals is required.}
\label{fig:ea-flowchart}
\end{figure}

Interestingly, GE can be used with no modification to optimize single attackers: when the maximum size of a group is set to $1$, the evolutionary core automatically stops using group manipulation operators, such as \texttt{addElementToGroup} and \texttt{removeElementFromGroup}. For the purpose of this work, we further modified the original GE available in $\mu$GP introducing a new mechanism for choosing which operators to use in the current generation, and a strategy for caching the results of past evaluations \cite{Beluz:GECCO15}. The flowchart of the GE algorithm used in this work is shown in Figure \ref{fig:ea-flowchart}.

\subsection{Internal solution representation}
\label{ssec:5method:encoding}

In the problem under study, we consider a model of \emph{strong colluding attackers} with full network knowledge and \emph{free range} of movement. In other words, we consider attacks carried out by multiple collaborating nodes which are connected via alternative communication links. In this context, a candidate solution represents a group of one or more malicious nodes, each one characterized by the following properties:

\begin{itemize}
\item the \emph{attack logic} to adopt (e.g., black hole, flooding); we assume this choice remains unchanged during an attack;
\item the \emph{movement model} (e.g., pedestrian, vehicle), which constrains the node to a corresponding map layer and mobility pattern;
\item the \emph{route} on any given map layer, defined via a list of POIs on that map layer; free-range strong attackers have full control when deciding their POIs.
\end{itemize}

The resulting structure of a candidate solution is thus quite complex, as each attacker can feature a variable number of POIs, and each group can have a variable number of attackers. An example of solutions produced by the evolutionary core is shown in Figure~\ref{fig:individuals}.

\begin{figure}[htb]
\centering
{\footnotesize
\begin{tabular}{l|l|l}
\texttt{Mov=vehicle} & \texttt{Mov=pedestrian} & \texttt{Mov=boat}\\
\texttt{Attack=black\_hole} & \texttt{Attack=flood} & \texttt{...}\\
\texttt{94,39} & \texttt{22,75}& \texttt{...}\\
\texttt{55,84} &\texttt{43,15} & \texttt{...}\\
\texttt{...} & \texttt{...} & \texttt{...}\\
\texttt{42,44} & \texttt{65,61} & \texttt{...}\\
\texttt{1,26} & \texttt{15,58} & \texttt{...}\\
\end{tabular}
}
\caption{Example of solution for the DTN attack problem, describing multiple attackers. Each attacker is characterized by its movement model (e.g. boat, pedestrian, vehicle), its attack logic (black hole or flood), and a series of POIs it will visit during the simulation, encoded as squares in a grid overlapped to the city map.}
\label{fig:individuals}
\end{figure}

In a classical EA, the individual would then be a (either fixed- or variable-size) composition of multiple attackers with different properties. When GE is applied instead, each individual models exactly a single attacker, while the organization of such malicious nodes is delegated to groups, with a final result structurally similar to the case where a single individual portrays several nodes. In any case, the genetic operators can act on each node, modifying its movement model, its attack logic, and adding/removing/replacing POIs in its path, in order to generate new candidate solutions. In experiments with a variable number of malicious nodes, the same operations can be performed on the blocks representing the nodes.

It is important to notice that the same POIs have different meanings depending on the node's movement: even if a vehicle and a pedestrian pass close to the same coordinates, they may reach them using different paths, causing distinct network disruptions along the way. Also, since most of the POIs are accessible by certain types of movement only (e.g., a point in open water cannot be reached by a pedestrian, nor one on land by a boat), for each type we overlap a \emph{grid} layer onto the city map layer, and define the path of an attacker of that type as a set of grid squares inside that grid. During the simulation, we then map each grid square to the map point \emph{closest} to the square; if a square contains more than one map point, the malicious node visits them all.

In particular, $\mu$GP exploits a user-defined description of solutions in an XML file. The external representation of candidate solutions is written to text files as they are evaluated, while their internal structure is stored to disk as XML files, that can be read in case the evolutionary process is resumed. 


\subsection{Fitness functions}
\label{ssec:5method:fitness}

Movement model, attack logic and POIs to visit are set at the level of each malicious node, but the objective is to maximize the global effectiveness of the attack: an optimal set of colluding attackers should lower the performance objectives of the network the most.
The effectiveness of an attack configuration is thus assessed as an evaluation of an urban network scenario (see next Section) 
performed by the network simulator. 
The outputs of such simulation (i.e., the fitness values, using the terminology of EC), are:
\begin{itemize}
\item ($f_1$) the data delivery rate (DDR), calculated as the percentage of messages originated \emph{only} from honest nodes, and which are delivered successfully;
\item ($f_2$) similarly, the average latency of message deliveries (in seconds).
\end{itemize} 
The two values are considered in lexicographic order, as we assign more importance to a reduction of the network's DDR rather than an increase of latency: optimization-wise, $f_1$ is to be minimized, while $f_2$ is to be maximized.

\section{Experiments}
\label{sec:exp}

This section first summarized the configuration settings for the experimental campaigns. Sections~\ref{sec:exp-DTN} and~\ref{sec:exp-users} quantify the parameters used to define an urban setting and its network nodes. Section~\ref{sec:exp-EA} summarizes the experimental parameters of the evolutionary core. Sections~\ref{sec:exp-results} and~\ref{sec:exp-discussion} give the numerical results and discuss their practical impact.

We make public the city maps, the experimental configurations, and detailed experimental results, at the URL:\\ \url{https://github.com/doinab/DTN-security}.

\subsection{DTN and the cities: San Francisco and Venice}
\label{sec:exp-DTN}

\begin{figure}[htb]
\begin{center}
\includegraphics[scale=0.37, clip, trim=2.5cm 1.3cm 2.5cm 2.1cm]{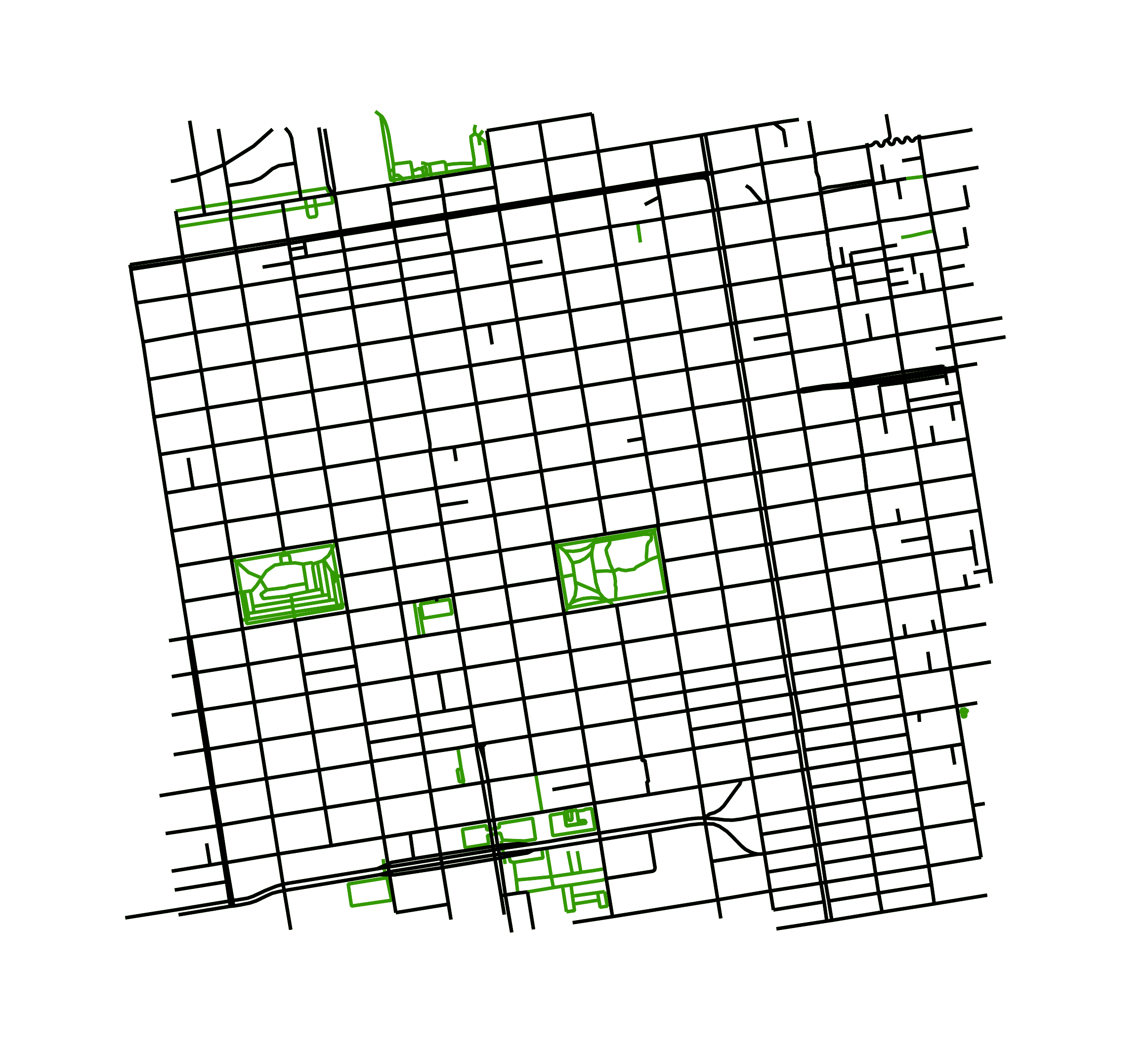}
\end{center}
\caption{A 5 km$^2$ area of downtown San Francisco, US, with a grid-based map topology of streets and the occasional park. The map has two overlapping layers, constraining the movement of vehicles and pedestrians: the vehicles are confined to the black streets, while the pedestrians may walk both the green and the black routes.}
\label{fig:map-SanFrancisco}
\end{figure}

\begin{figure}[htb]
\begin{center}
\includegraphics[scale=0.4, clip, trim=2.6cm 6.7cm 2.6cm 6cm]{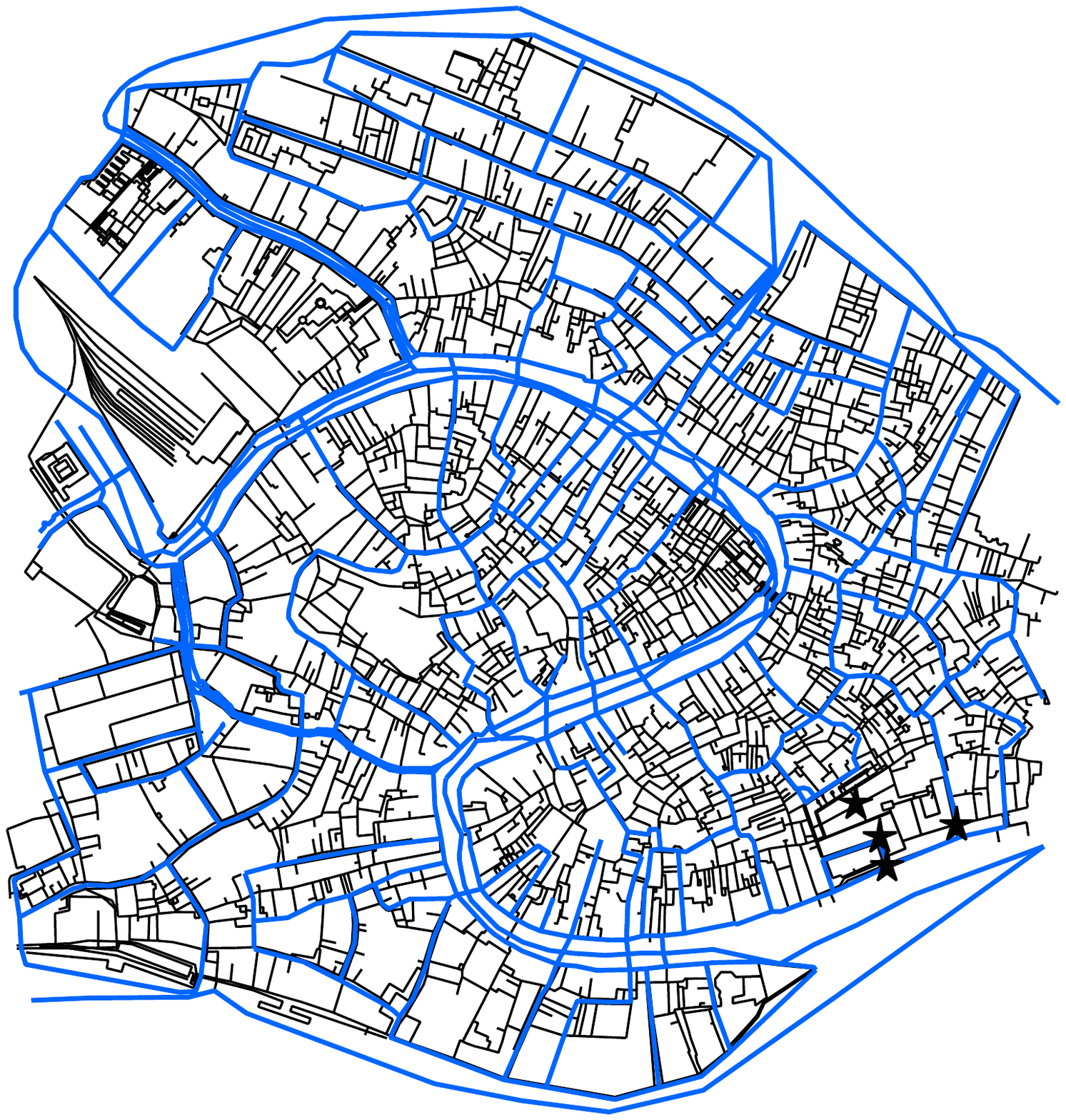}
\end{center}
\caption{A 5 km$^2$ area of downtown Venice, IT, with an irregular map topology of pedestrian pathways (the black map layer) and waterways (the blue layer). Marked with stars are special POIs in the city's touristic center.}
\label{fig:map-Venice}
\end{figure}

To validate our methodology, we simulate two realistic, large-scale city environments, each composed of a map and a large set of honest, randomly moving network nodes of certain types relevant to a given city. Figures~\ref{fig:map-SanFrancisco} and~\ref{fig:map-Venice} show the basic maps of the two urban environments in our experimental scenarios, namely San Francisco and Venice. These cities differ in terms of map topology: while the area of San Francisco has a regular grid structure of routes, the area of Venice has a complex, hierarchical, irregular structure of main and secondary waterways travelled by boats, with pedestrians confined to inner walkways (some along waterways) and bridges. The Venice map has an additional feature for added realism: on both map layers, a small number of the map POIs mark the touristic center, and have a higher probability to be chosen as the next destination by the honest nodes. Table~\ref{tab:sim-settings-city} quantifies the maps and map layers in terms of size, number of distinct map points, route segments, and number of nodes. 

\begin{table*}[ht!]
\centering
\caption{Network parameters: city maps}
\label{tab:sim-settings-city}
{\footnotesize
\begin{tabular}{p{2.5cm} r@{\hspace{4mm}} p{7.5cm}}
\toprule
	\multirow{5}{*}{\parbox{2.5cm}{\textbf{San Francisco:}}}	&
	size:	&
	2416 m $\times$ 2253 m \\
	&
	map layers:	&
	$L_P$ (pedestrian walkways), $L_S$ (streets)	\\
	&
	no.\ of route segments:	&
	1728 in $L_P$, 1305 in $L_S$	\\
	&
	no.\ of map points:	&
	1210 in $L_P$, 883 in $L_S$	\\
	&
	network size:	&
	150 pedestrians (constrained to $L_P$), 
	50 cars (constrained to $L_S$) \\
\midrule
	\multirow{5}{*}{\parbox{2.5cm}{\textbf{Venice:}}}	&
	size:	&
	2210 m $\times$ 2340 m \\
	&
	map layers:	&
	$L_P$ (pedestrian walkways), $L_W$ (waterways)	\\
	&
	no.\ of line segments:	&
	7983 in $L_P$, 1497 in $L_W$	\\
	&
	no.\ of map points:	&
	6910 in $L_P$, 1354 in $L_W$	\\
	&
	network size:	&
	150 pedestrians (constrained to $L_P$), 
	50 boats (constrained to $L_W$) \\
\bottomrule
\end{tabular}
}
\end{table*}

\subsection{Network simulation and network nodes}
\label{sec:exp-users}

In both cities we configured $N=200$ moving network nodes, divided in two types: pedestrians ($75\%$) and vehicles ($25\%$). For San Francisco, the vehicles consist of motorized cars; in Venice, the waterways serve as routes for motorized or unmotorized boats. Pedestrians are modelled as carrying communication devices with relatively limited capabilities: a Bluetooth communication interface with a range of 15m and low bandwidth. Vehicles are awarded more communication capabilities: besides a Bluetooth interface (which allows communication events to take place between any pedestrian and any vehicle), a vehicle also has a high-speed, longer-range network interface allowing vehicle-to-vehicle communication. 

Each simulation of a DTN in The ONE is stochastic. The nodes are initially placed randomly on their map layer, and a 1,000-second warm-up simulation period is allowed before the experiment starts, for the nodes to settle on the ``natural'' preferred routes in the city. The next destination POI is also chosen randomly. Due to this, to smoothen the fitness landscape and reduce the effect of the random seed of each simulation on the evaluation of solution, we execute each network simulation 10 times, initialized with different random seeds, and report as fitness values the average DDR and latency over the $10$ available repetitions.

The movement model of all nodes follows the general randomized pattern summarized in Section~\ref{sec:background-movement-models}. A subset of these $200$ nodes is assigned a malicious behaviour. For an \emph{honest} node, the set of POIs is simply the entire set of map points located on the node's relevant map layer. The node randomly chooses any destination point from that map layer, travels there at a certain speed on the shortest path, pauses for an interval, and repeats the process. The configuration for the nodes' speed and pause interval is given by Table~\ref{tab:sim-settings-movement}. For an \emph{attacker}, the set of POIs is is a subset of the map points of the relevant map layer, and is evolved by the evolutionary core as part of each solution (as described in Section~\ref{ssec:5method:encoding}). The movement model of an (e.g., pedestrian) attacker then only differs from that of an honest pedestrian in that the attacker's next destination point is randomly chosen from the evolved set of POIs, rather than the entire map layer.

\begin{table*}[ht!]
\centering
\caption{Network parameters: movement models}
\label{tab:sim-settings-movement}
{\footnotesize
\begin{tabular}{p{2.5cm} r@{\hspace{4mm}} p{7.5cm}}
\toprule
	\multirow{6}{*}{\parbox{2.5cm}{\textbf{Movement model\\ for nodes in all cities}}}	&
	next point:	&
	chosen randomly from a map layer \\
	&
	path choice:	&
	shortest path on the map layer to the next point \\
	&
	pedestrian speed:	&
	$[0.5 \ldots 1.5]$ m/s \\
	&
	boat speed:	&
	$[1.0 \ldots 5.0]$ m/s \\
	&
	car speed:	&
	$[2.7 \ldots 13.9]$ m/s \\
	&
	pause interval for all:	&
	$[0 \ldots 120]$ s at each destination point \\	
\bottomrule
\end{tabular}
}
\end{table*}

\begin{table*}[ht!]
\centering
\caption{Network parameters: simulation and node communication settings}
\label{tab:sim-settings-comm}
{\footnotesize
\begin{tabular}{p{2cm} r@{\hspace{4mm}} p{8cm}}
\toprule
	\multirow{2}{*}{\parbox{2cm}{\textbf{Simulation\\ settings}}}	&
	simulation time:	&
	5 h \\
	&
	DTN simulator:	&
	The ONE~\cite{ONE} \\
\midrule
	\multirow{4}{*}{\parbox{2cm}{\textbf{Message\\ settings}}}	&
	message issued:	&
	every 30 s (by an honest node), every 3 s (by a flooder) \\
	&
	message size:	&
	10 kB (issued by an honest node), 100 kB (issued by a flooder) \\
	&
	message buffer:	&
	5 MB (for pedestrian nodes), 50 MB (for car and boat nodes)\\
	&
	message TTL:	&
	5 h \\
\midrule
	\multirow{4}{*}{\parbox{2cm}{\textbf{Node\\ communication \\interfaces}}}	&
	Bluetooth:	&
	range 15 m, speed 250 kBps \\
	&
	High-speed:	&
	range 100 m, speed 10 MBps \\
	&
	pedestrians use:	&
	Bluetooth \\ 
	&
	cars and boats use:	&
	Bluetooth and High-speed \\ 
\bottomrule
\end{tabular}
}
\end{table*}

Honest nodes periodically inject new messages to be routed by the network; the rate of message injection among all honest nodes is set at one message every 30 seconds, such that the network routes 120 honest messages per hour. The honest node to inject the next message in the network is chosen randomly. The malicious nodes run one of the two attack logics described in Section~\ref{sec:background-security}. 

A black hole attacker does not inject any additional messages in the network. On the other hand, when an attacker executes a flood, the parameters are chosen to obtain a ``heavy'' flood of messages: (1) a flooding node injects messages in the network at 10 times the frequency of message injection from an honest node, and (2) the messages injected by a flooder are 10 times as large as regular messages. Table~\ref{tab:sim-settings-comm} summarizes these communication parameters, together with the settings regarding the sizes of the nodes' message buffers, and the Time To Leave (TTL), which limits the amount of time that a message is allowed to be stored in a node's buffer without being forwarded --- we set TTL to be large, and equal to the length of an experiment: 5 hours (simulated time).

\begin{figure*}[ht!]
	\centering
	\includegraphics[scale=0.5]{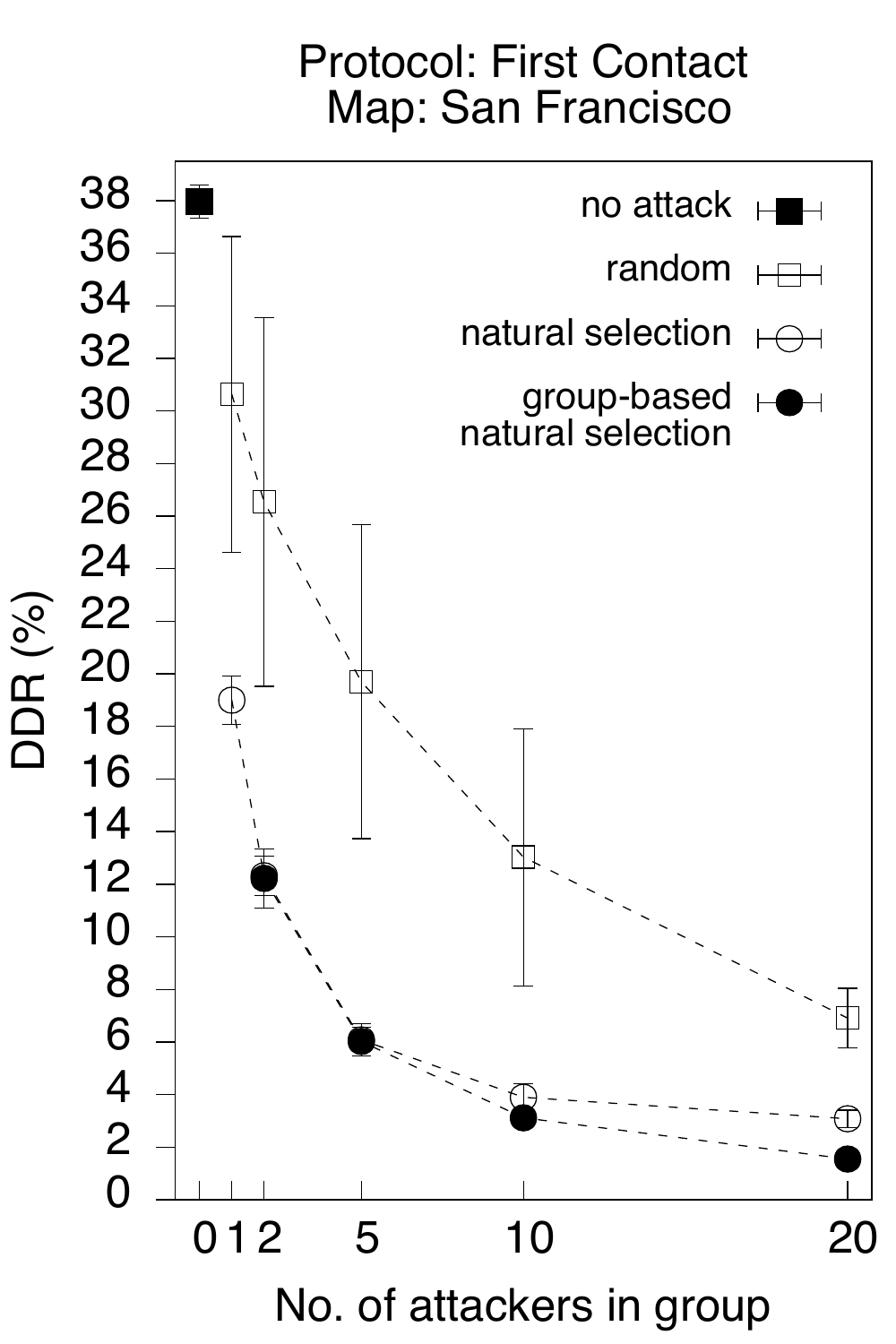}
	\hspace{1cm}
	\includegraphics[scale=0.5]{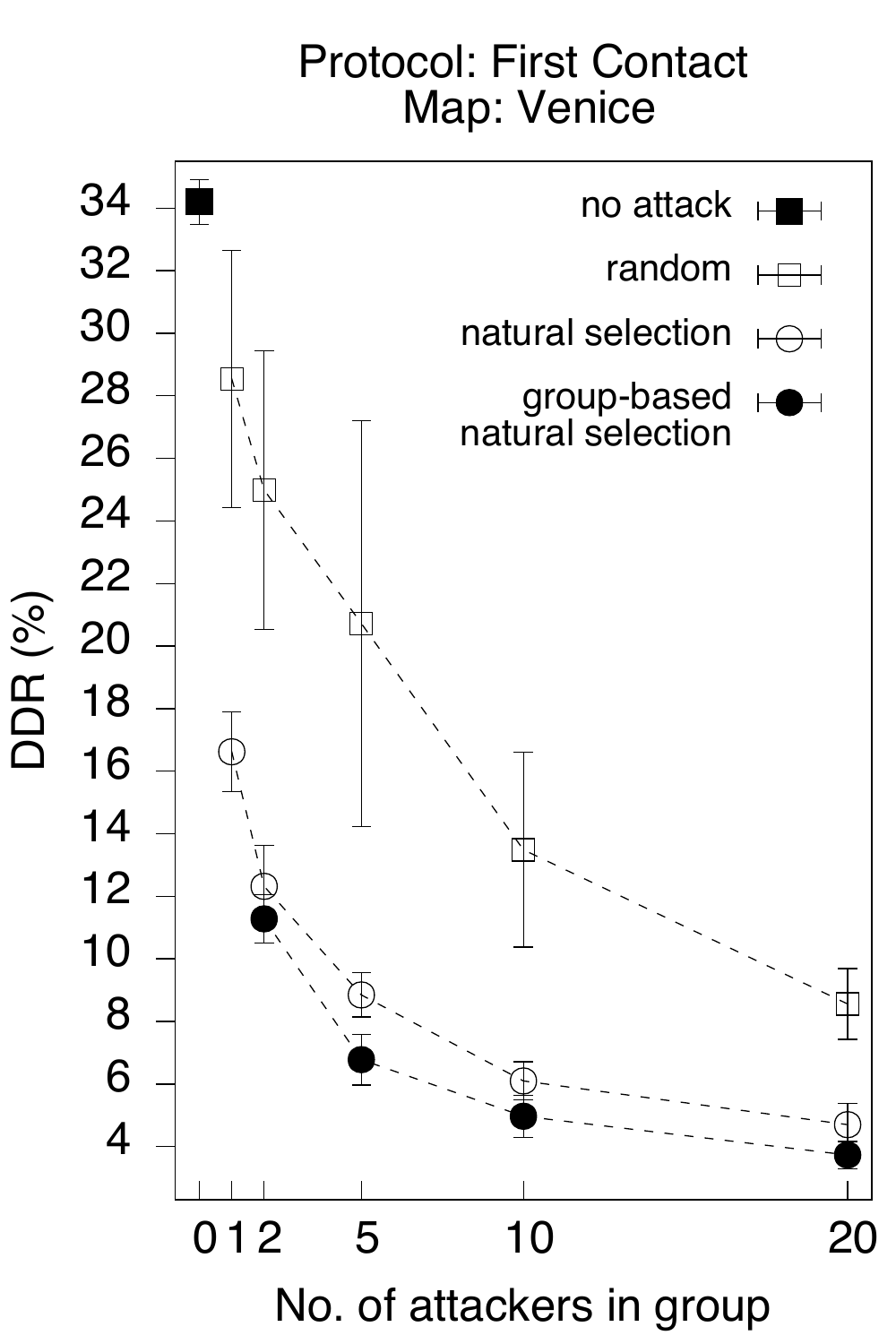}
	\caption{Comparative results: data delivery rate (DDR) from three experimental campaigns using group evolution, a classical EA, and random testing. The DDR obtained with a random test is shown as the mean and standard deviation among the mean DDR of 150 randomly generated groups, each group simulated 10 times with different random seeds. The DDR obtained with any evolutionary experiment is a single group of top fitness, and is shown as the mean and 95\% confidence interval among 5 simulation repetitions of that group with different random seeds.}
	\label{fig:overview-results}
\end{figure*}

\subsection{Evolutionary parameters}
\label{sec:exp-EA}

During all the experiments, $\mu$GP has been configured with the parameters reported in Table~\ref{tab:ugp_par}. The operators chosen for the evolution are:
\begin{itemize}
 \item \texttt{onePointImpreciseCrossover}: one-point crossover between two individuals;
 \item \texttt{twoPointImpreciseCrossover}: crossover with with two cut points;
 \item \texttt{singleParameterAlterationMutation}: mutate a single coordinate of a POI, the movement model or the attack logic;
 \item \texttt{insertionMutation}: add a new random POI;
 \item \texttt{removalMutation}: remove a randomly selected POI;
 \item \texttt{replacementMutation}: replace a POI with a randomly generated one.
\end{itemize}
The activation probabilities of all the operators in the population are self-adapted during the run; \cite{Beluz:GECCO15} enables to efficiently alternate phases where individuals are optimized, with phases where groups are optimized.
Self adapting the size of the tournament for selecting parents ($\tau$) enables to optimize the selective pressure; while self adapting the strength of the mutation operators ($\sigma$) enables to balance between exploration and exploitation. Every time a mutation is performed, it is executed again on the same individual if a randomly generated number in $(0,1)$ is lower than the current value of $\sigma$. For high values of $\sigma$, the algorithm will tend to generate new solutions that are very different from their parents, thus favoring exploration; for low values of $\sigma$, the differences between parents and offspring will be smaller, thus entering a phase of exploitation.

\begin{table}[ht!]
\caption{$\mu$GP experimental settings}
\label{tab:ugp_par}
{\footnotesize
\begin{center}
\begin{tabular}{l@{\hspace{4mm}} l@{\hspace{4mm}} c}
\toprule
Parameter & Description & Value \\
\midrule
$\tau$ & size of the tournament selection & 1.0$\div$4.0 \\
$\sigma$ & initial strength of the mutation operators & 0.9 \\
$\alpha$ & inertia of the self-adapting mechanisms & 0.9 \\
$\S$ & stagnation threshold (in generations) & 50 \\
\hline
\multicolumn{3}{c}{Classical EA}\\
\hline
$\mu$ & individual population size & 30 \\
$\lambda$ & operators (genetic) applied at every step & 20 \\
\hline
\multicolumn{3}{c}{Group Evolution}\\
\hline
$\mu_{group}$ & group population size & 30 \\
$\nu_{individual}$ & initial individual population size & 50 \\
$\lambda$ & operators (genetic or groups) applied at every step & 20 \\
\bottomrule
\end{tabular}
\end{center}
}
\end{table}

During the GE experiments, new operators are added in order to manipulate groups, namely:
\begin{itemize}
 \item \texttt{groupRandomInsertionMutation}: add a random individual to a group;
 \item \texttt{groupRandomRemovalMutation}: remove a random individual from a group;
 \item \texttt{groupBalancedCrossover}: crossover that moves the same number of individuals between two groups;
 \item \texttt{groupUnbalancedCrossover}: same as above, with no guarantee of moving the same number of individuals;
 \item \texttt{groupUnionIntersection}: returns the union and the intersection of two groups;
 \item \texttt{groupDreamTeam}: creates a group with some of the best individuals currently in the population.
\end{itemize}
As previously detailed, the number of individuals in the GE paradigm is regulated by the number of groups in the current population, $\mu_{group}$. Individuals are removed only when they are no longer included in any group. The number of individuals generated at the beginning of the execution, number that can be later exceeded or reduced, is controlled by parameter $\nu_{individual}$. Further information on parameters and operators in $\mu$GP can be found in \cite{Sanchez2011}. 

\subsection{Experimental campaigns and results}
\label{sec:exp-results}

Given the settings described in Sections~\ref{sec:exp-DTN}--\ref{sec:exp-EA}, which are common to all experiments, an experiment will be uniquely identified by the following two parameters:
\begin{description}
	\item[City] (i.e., San Francisco or Venice)
	\item[Number of attackers] i.e., the size of the attack group $k$, in the range $1\le k\le N$. We delineate five practically interesting ranges:
		\begin{itemize}
			\item $k=1$, i.e., a single attacker;
			\item $k=2$, i.e., a pair of attackers;
			\item $k\in[1\ldots 5]$;
			\item $k\in[6\ldots 10]$;
			\item $k\in[11\ldots 20]$, i.e., a group which can reach 10\% of the overall network size of $N=200$ nodes. 
		\end{itemize}
\end{description}

The ranges for $k$ thus come in two categories: a fixed group size (when $k=1$ or $k=2$), or a variable group size (for larger $k$). In the latter case, intuitively, the expectation is that the evolutionary algorithm will maximize the group size in the process of optimizing the fitness functions. 

To assess the comparative performance of our method based on Group Evolution, we ran the following three experimental campaigns, for each experimental setting (City $\times$ Number of attackers):
\begin{itemize}
	\item A GE-based experimental campaign (for the settings where $k>1$);
	\item An experimental campaign based on the classical, non-GE EA applied to the same problem in prior literature~\cite{Bucur:EvoApp15};
	\item The testing of groups of a sample of 150 purely randomly generated groups of attackers.
\end{itemize}
For each experimental setting, the GE and non-GE experimental campaigns consist of 5 experiment repetitions, initialized with different random seeds.

The results of the experimental campaigns are summarized in Figure~\ref{fig:overview-results}, separately per city. The first fitness function, $f_1$, measuring the global data delivery rate (DDR) in the network, quantifies the decreasing network performance with an increasing size of the attack group, $k$. In the figures, the data points are presented for the set $k\in \{1,2,5,10,20\}$, as the evolutionary algorithms found that the lowest fitness is achieved when the size of the attacker group is maximum. For comparison, the figures also include a data point showing the fitness function with no attack present (i.e., all $N=200$ nodes in the network are honest).

As seen in Figure~\ref{fig:overview-results}, while the First Contact protocol has only moderate data delivery in these complex urban settings even in the absence of attacks, both evolutionary algorithms significantly outperform random testing, and GE was found to be advantageous in all the experimental settings. Moreover, a similar trend for the performance of the First Contact protocol was obtained between the two cities: a single attacker was found sufficient to lower the data delivery to half that of the no-attack setting, and, with a group of 20 attackers, the data delivery in the network was found to drop close to zero.

To further confirm this trend, we performed a thorough analysis \cite{lehmann2005testing} of the numerical results. The analysis was conducted as follows: for each city map and attack group size $k>1$ (when $k=1$ a group cannot be defined), first we aggregated the 150 lowest DDR values obtained by GE and non-GE experiments, over the 5 available repetitions for each of the two algorithms. We then performed pairwise comparisons between the two algorithms, and w.r.t. the DDR of the 150 randomly sampled attack groups used as baseline. For each pairwise comparison, we initially verify the normality of the two distributions with the Shapiro-Wilk test; if both samples are normally distributed, we then test the homogeneity of their variances (homoscedasticity) with an F-test. If variances are equal, we compare the two distributions by means of the Student \emph{t}-test, otherwise we adopt Welch's \emph{t}-test variant. More specifically, we first test the null-hypothesis of equal distributions (i.e. the two algorithms under comparison are statistically equivalent from an optimization point of view); then, we test the null-hypothesis that the fitness values obtained with one of the two algorithms, taken as reference, are statistically smaller than those obtained by the other algorithm. In case of non-normal distributions, we instead test the null-hypotheses by means of the non-parametric Wilcoxon Rank-Sum test. In all the tests, we consider a confidence level of $0.95$ ($\alpha=0.05$).

The statistical analysis is summarized in Table~\ref{tab:stats}. The analysis confirms that GE and non-GE EA statistically outperform random sampling, and GE outperforms non-GE in all cases.

\begin{table}[ht!]
\centering
\caption{Summary of the statistical analysis (see the main text for details). 
Each column (labeled as X/Y) shows the pairwise comparison between the results obtained 
by algorithm X and Y, with X taken as reference. The symbol '+' indicates that X 
statistically outperforms Y, i.e. it obtains attacker groups with lower network performance (DDR).}
\label{tab:stats}
\footnotesize
\begin{tabular}{|c|c|c|c|}
\hline
\multicolumn{4}{|c|}{San Francisco}\\
\hline
No. attackers & GE/non-GE & GE/Random & non-GE/Random\\
\hline
2 & + & + & + \\
1-5 & + & + & + \\
6-10 & + & + & + \\
11-20 & + & + & + \\
\hline
\end{tabular}
\\
\begin{tabular}{|c|c|c|c|}
\hline
\multicolumn{4}{|c|}{Venice}\\
\hline
No. attackers & GE/non-GE & GE/Random & non-GE/Random\\
\hline
2 & + & + & + \\
1-5 & + & + & + \\
6-10 & + & + & + \\
11-20 & + & + & + \\
\hline
\end{tabular}
\end{table}

\subsection{Runtimes and discussion of results}
\label{sec:exp-discussion}

\begin{figure}[ht]
	\includegraphics[scale=0.5]{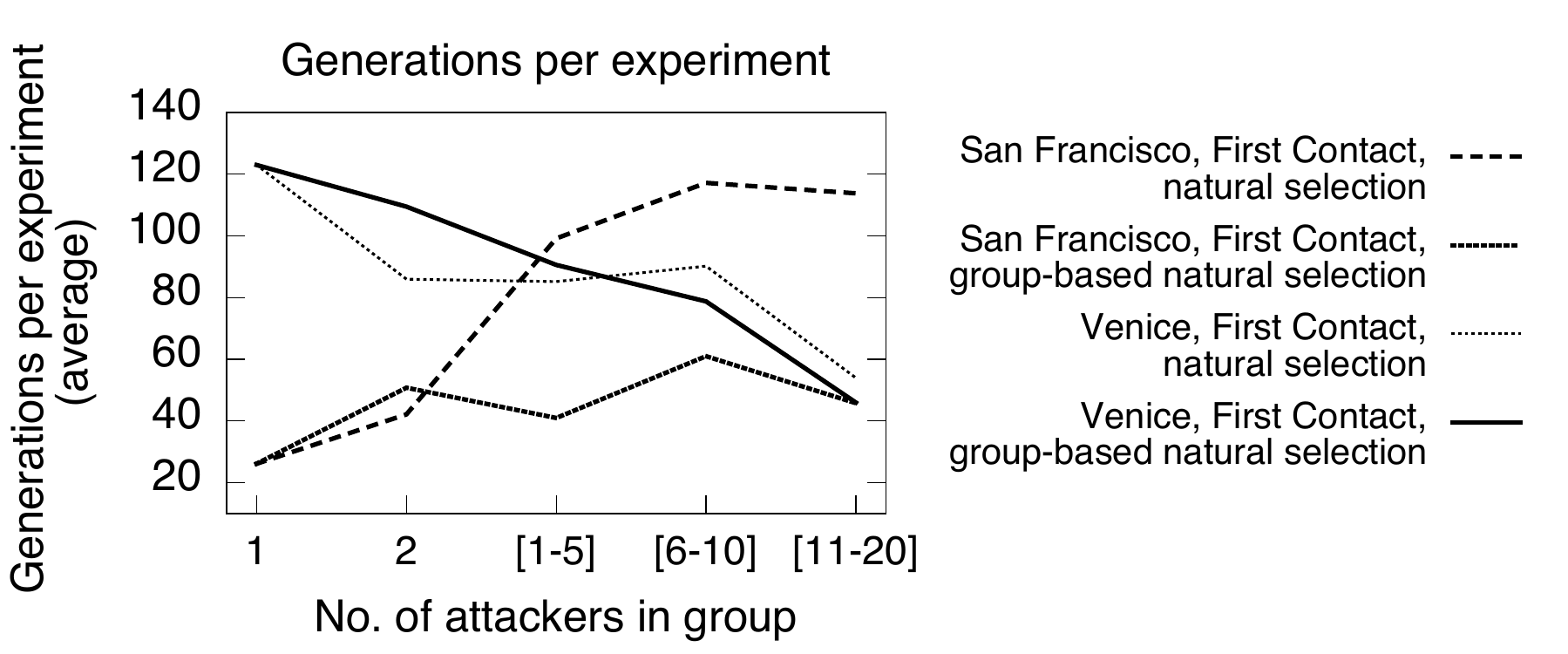}
	\includegraphics[scale=0.5]{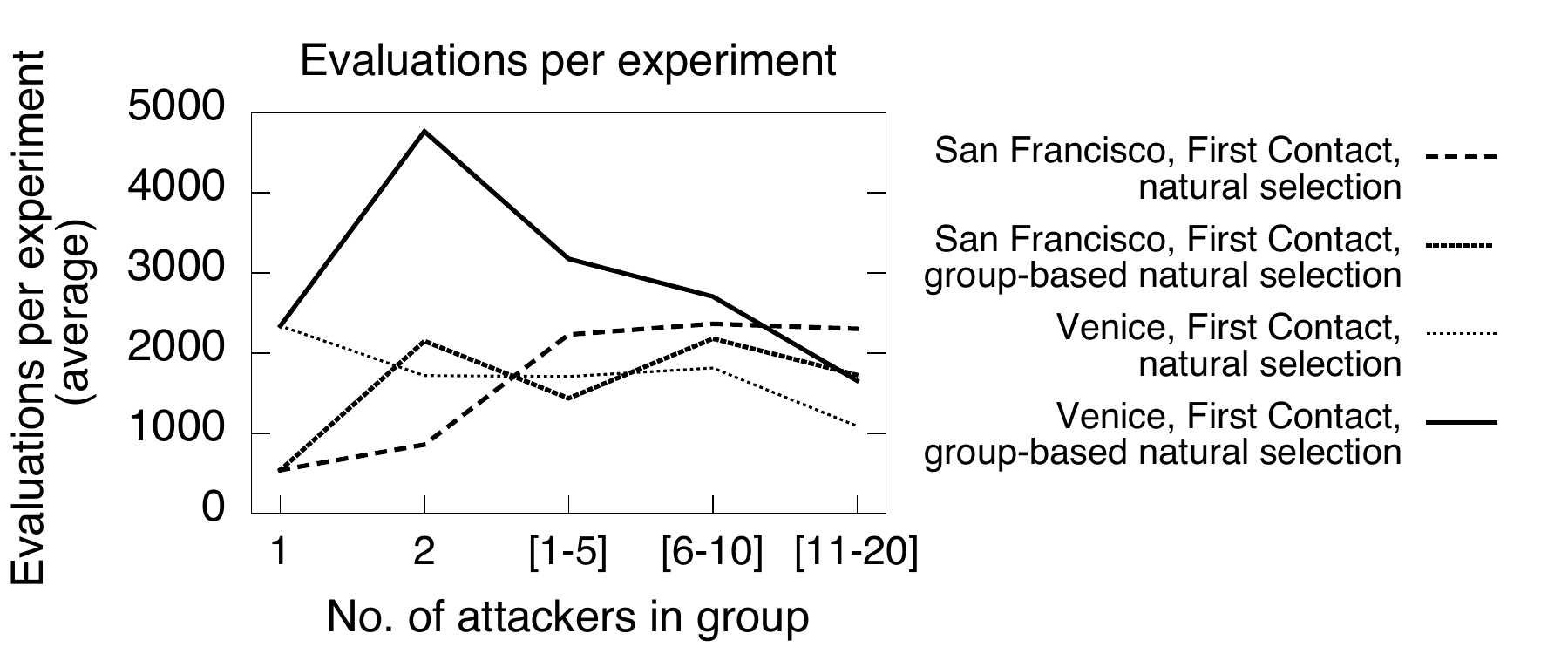}
	\includegraphics[scale=0.5]{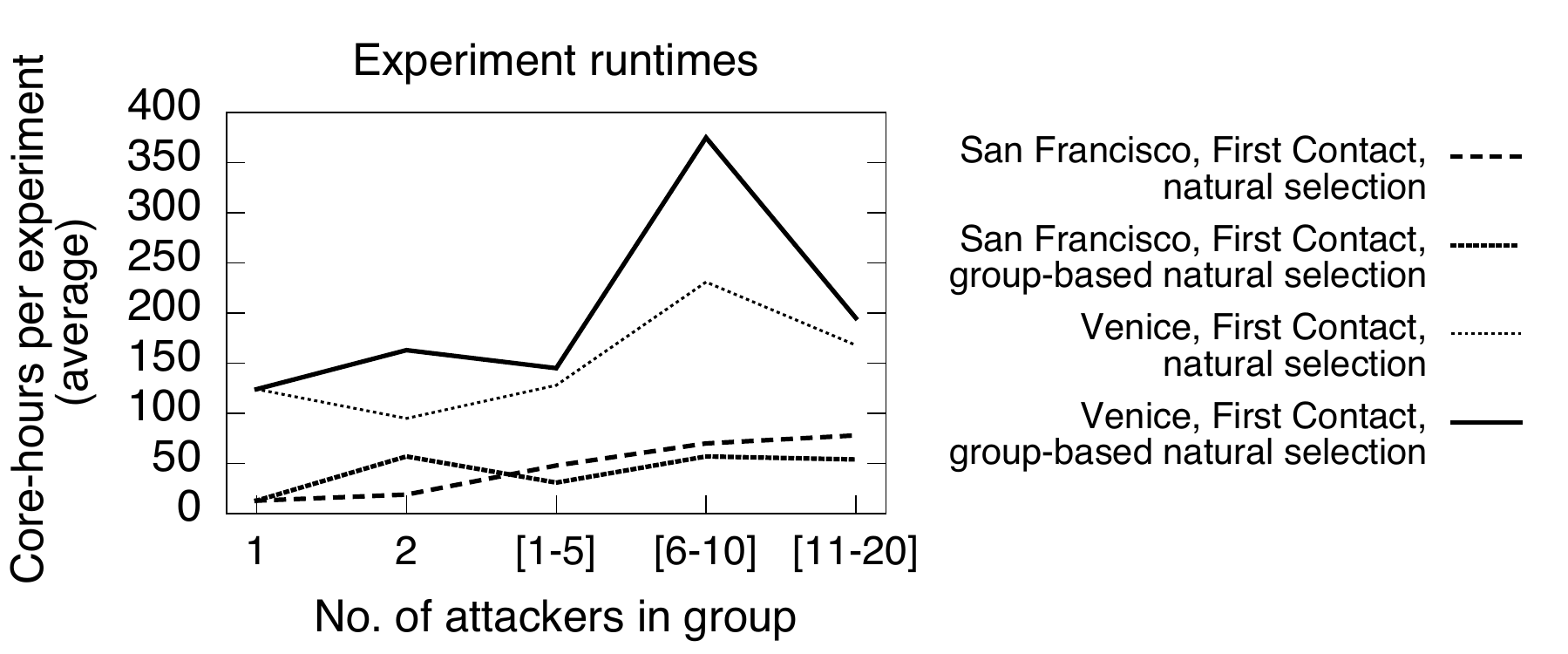}
	\caption{Computational overhead of non-GE and GE experiments: the number of evolutionary generations, the number of solution evaluations, and the wall-clock runtime per experimental campaign, shown as the average among the 5 experiments configured with different random seeds, per experimental setting.}
	\label{fig:runtimes}
\end{figure}

In Figure~\ref{fig:runtimes} we show the computational cost of running the evolutionary campaigns, each data point the average of 5 repetitions of an experiment with different seeds. The runtime is shown in core-hours, over computing cores of at least 1.6 GHz (since we ran the experiments on a number of machines, there was variation among the computational power allowed among experiments).

\begin{figure*}[ht!]
	\centering
	\includegraphics[scale=0.5]{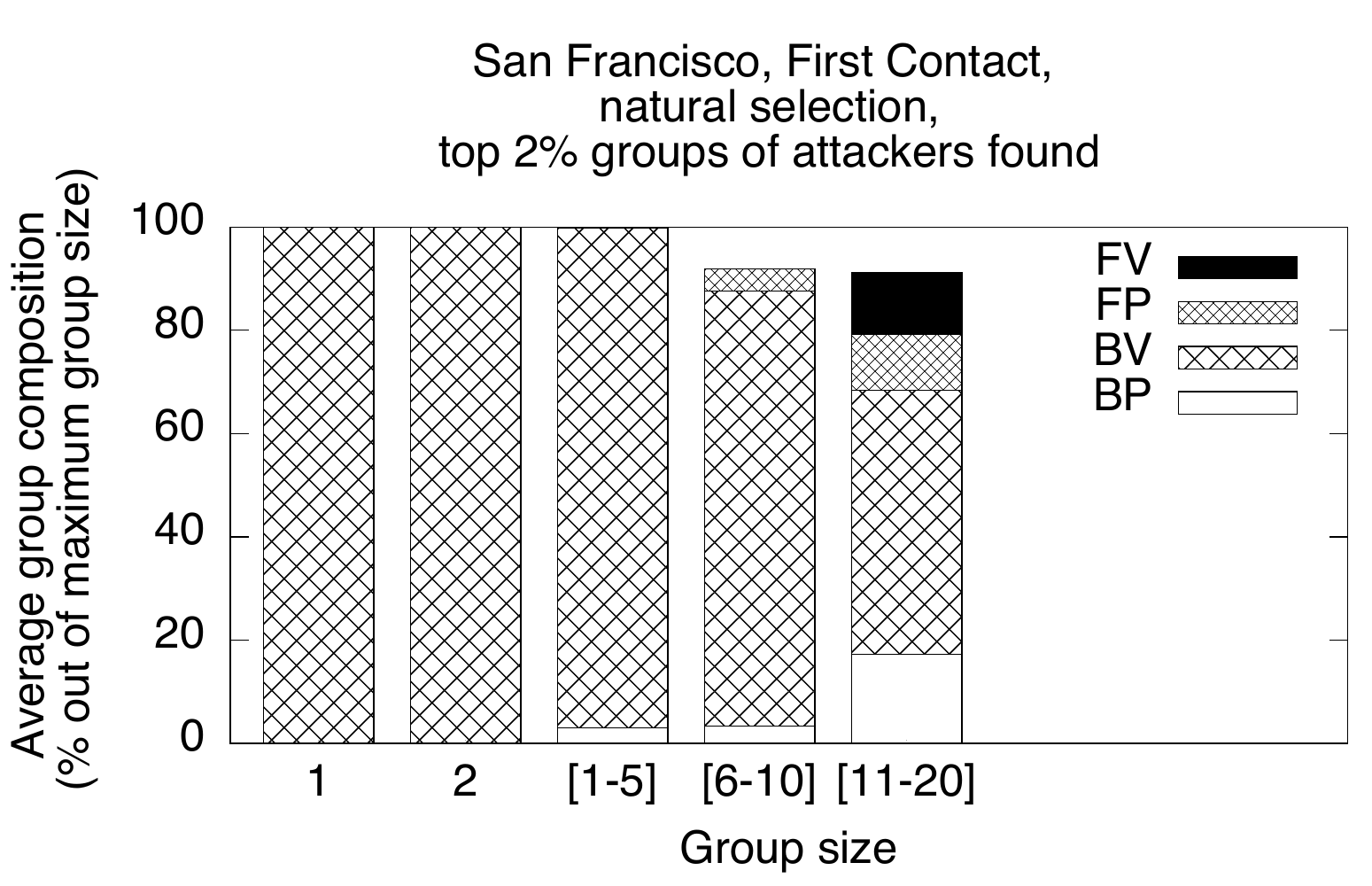}
	\hspace{1cm}
	\includegraphics[scale=0.5]{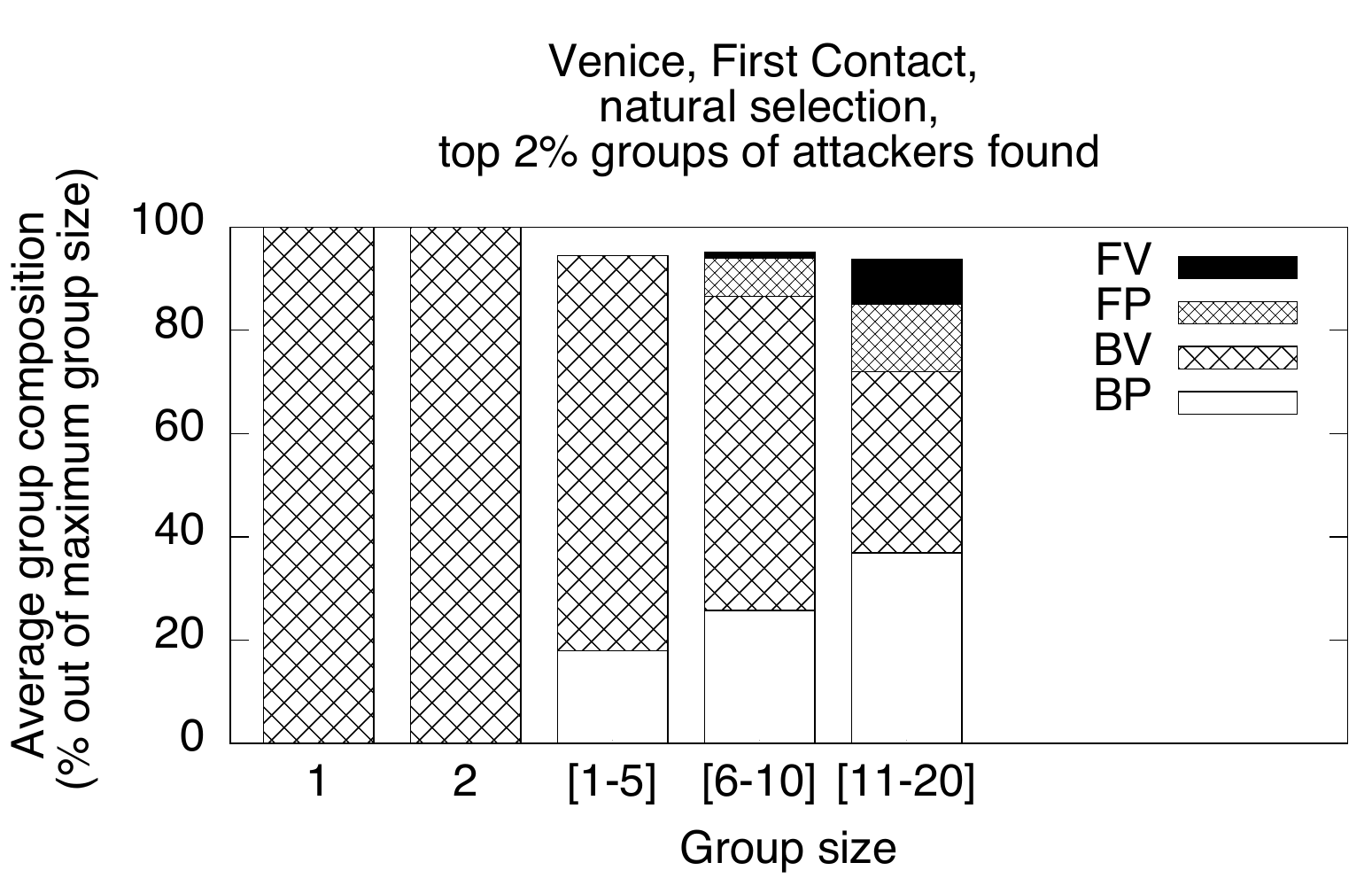}

	\includegraphics[scale=0.5]{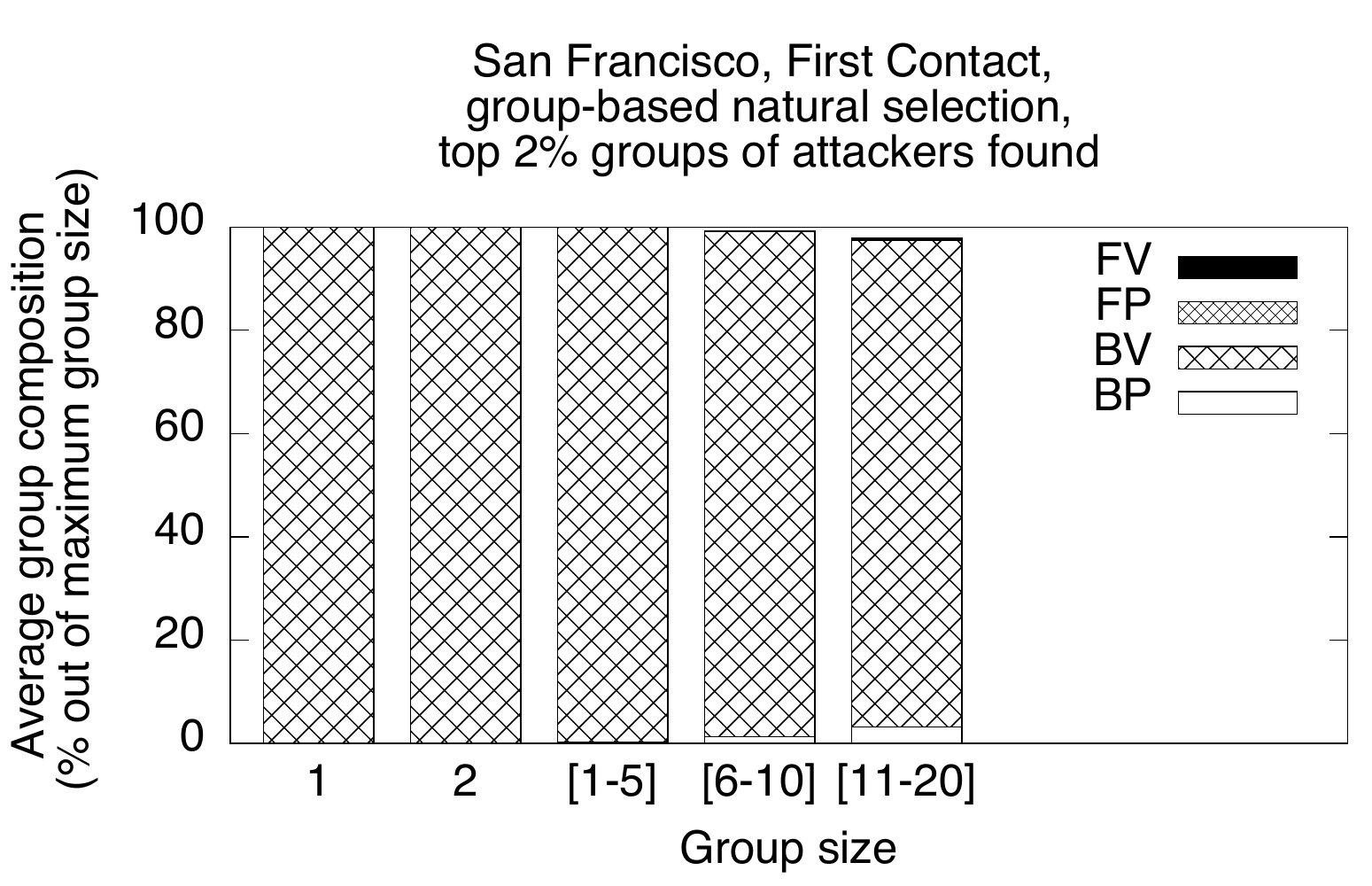}
	\hspace{1cm}
	\includegraphics[scale=0.5]{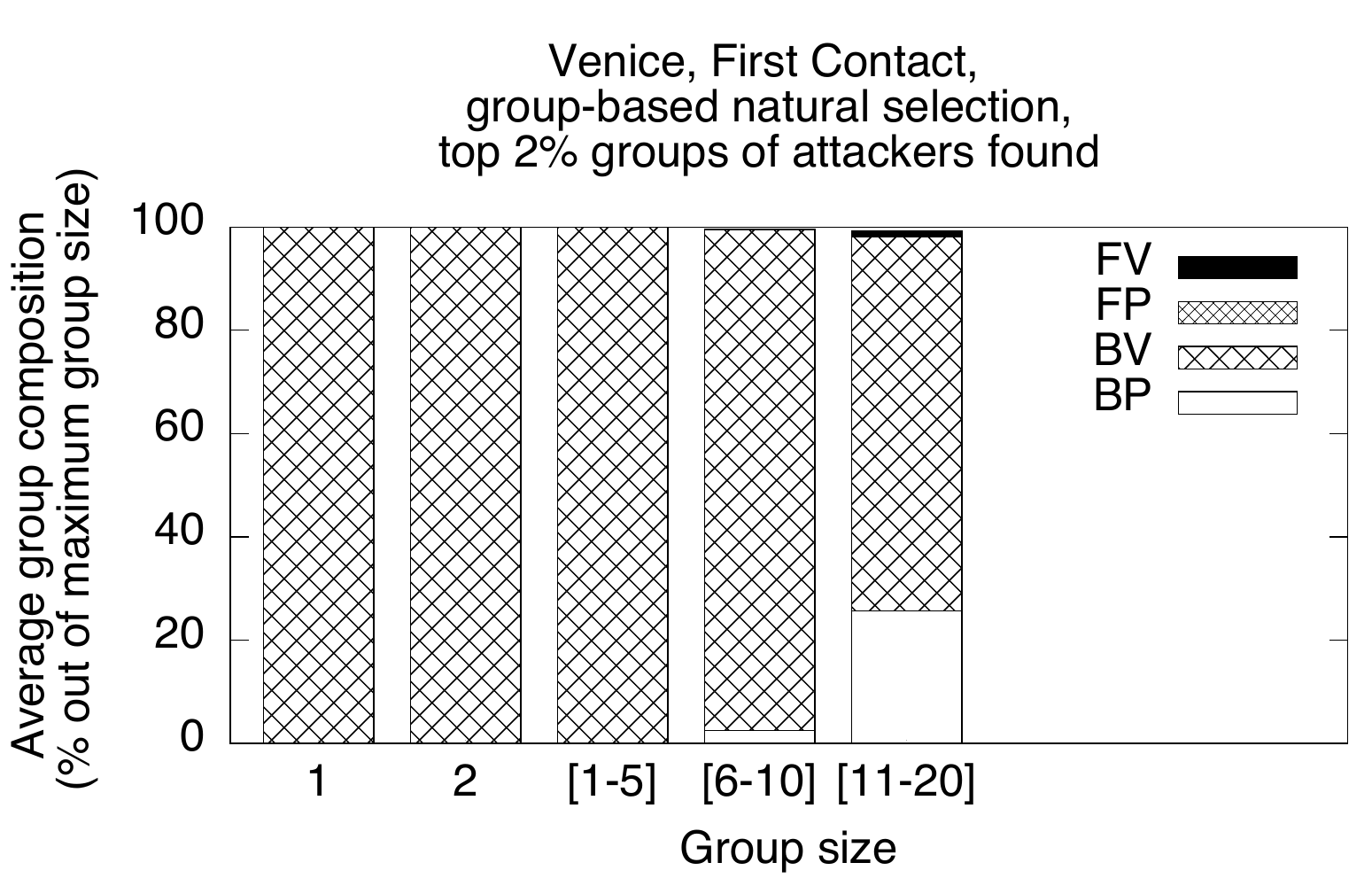}
	\caption{The average group size and group composition among all top solutions (whose fitness is within 2\% of the best fitness). The group composition is shown in terms of the percentage of flooding vehicles (FV), flooding pedestrians (FP), black hole vehicles (BV), and black hole pedestrians (BP) among these top solutions.}
	\label{fig:group-composition}
\end{figure*}

More interestingly, the reason why the GE algorithm shows an advantage over a classical EA is seen in Figure~\ref{fig:group-composition}. The figure first selects all the ``top'' attacker groups obtained by the GE and non-GE experiments, i.e., those groups which lower network performance to within 2\% of the absolute best fitness $f_1$ found by that algorithm. We observed that the number of these top groups falls between $300$ and $6,000$ (depending on the experimental setting), all of which can be considered successful attacks. The average composition of this large group sample is analyzed in terms of how often a type of attacker, i.e., a movement model (that of a pedestrian, or that of a vehicle) and attack logic (black hole or flooding) appears in a top group. 

The best single attacker found by both the GE and non-GE algorithms is a black hole vehicle: indeed, 100\% of the top groups found by both algorithms consist of an attacker of this type. The same is true for pairs of attackers. 

For larger group sizes, the two algorithms show a difference of results. When $k\in \left[ 11\ldots 20\right]$, the non-GE algorithm found an overall lower top fitness while obtaining top groups for which: (1) the average group size does not saturate (it reaches an average of 18.23 out of 20, for the San Francisco setting), and (2) the average group composition is a mix of attacker types, with only 50\% of the attackers in the top groups matching the type of the best single attacker (a black hole vehicle). 

On the other hand, the GE algorithm likely outperformed the non-GE in terms of best fitness found due to the fact that it both maximized the average top group size, and optimized the average top attacker type, demonstrating that homogeneous groups of black holes are advantageous to exploit the vulnerabilities in the design of the First Contact protocol, and that faster black hole attackers also have a clear advantage. 


\section{Conclusions}
\label{sec:concl}
In this paper we proposed a heuristic methodology to assess the robustness of \emph{First Contact}, one of the main routing protocols used in Delay-Tolerant Networks. To exploit possible weaknesses of the network, we considered the worst-case scenario of an attack carried out by a coordinated group of agents with full network knowledge. The methodology is based on an evolutionary algorithm using  a cooperative co-evolution scheme called \emph{Group Evolution} recently introduced in the literature and here extended for our purposes. The method is able to optimize groups (either homogeneous or not) of malicious nodes whose behaviour does not comply with the legitimate routing protocol used by honest nodes in the network.

We performed an extensive experimental campaign over medium-sized (i.e., 200 nodes) realistic urban networks running on two different cities with radically different map topologies (San Francisco and Venice). We assessed the scalability of the approach by evaluating single attackers as well as groups of up to $10\%$ malicious nodes. Moreover, we compared results obtained with random sampling and a more classical evolutionary algorithm.

In all our experiments, the two evolutionary methods clearly outperformed random sampling, consistently finding groups of attackers that produced a larger network damage (reduced data delivery rate and increased latency). The additional advantage brought by Group Evolution resulted in an improved attack effect, optimized group compositions, and more effective movement models.

Overall, the contribution of this work is twofold: on one hand, we proposed an efficient alternative to random sampling, that is currently one of the most used approaches for assessing network robustness; on the other hand, we showed an example problem that can be naturally described in terms of cooperative co-evolution and for which such an evolutionary scheme is clearly beneficial. 

This work represents then one of the few attempts at finding applications of cooperative co-evolution beyond the typical domain of swarm robotics. In future research, we will seek to extend this approach to different networking applications and, possibly, new unexplored domains.


\section*{References}
\balance
\bibliographystyle{model1-num-names}
\bibliography{man}

\end{document}